\documentclass[10pt,twocolumn,letterpaper]{article}

\usepackage[pagenumbers]{cvpr} 

\usepackage{graphicx}
\usepackage{amsmath,mathrsfs}
\usepackage{amssymb}
\usepackage{booktabs}
\usepackage{multirow}
\usepackage[table,xcdraw]{xcolor}
\usepackage{bbding}
\usepackage{stfloats}

\usepackage[pagebackref,breaklinks,colorlinks]{hyperref}

\usepackage[capitalize]{cleveref}
\crefname{section}{Sec.}{Secs.}
\Crefname{section}{Section}{Sections}
\Crefname{table}{Table}{Tables}
\crefname{table}{Tab.}{Tabs.}

\def\cvprPaperID{9521} 
\def\confName{CVPR}
\def\confYear{2023}

\begin{document}

\title{Boosting Low-Data Instance Segmentation by  Unsupervised Pre-training with Saliency Prompt}

\author{
    Hao Li$^{a,c}$, Dingwen Zhang$^a$, Nian Liu$^b$, Lechao Cheng$^c$,Yalun Dai$^d$, Chao Zhang$^e$\\
    Xinggang Wang$^f$, Junwei Han$^a$\\
    $^a$ Brain and Artificial Intelligence Lab, Northwestern Polytechnical University\\
    $^b$ Inception Institute of Artificial Intelligence
    $^c$  Zhejiang Lab\\
    $^d$  University of Chinese Academy of Sciences
    $^e$  NetEase
    $^f$  Huazhong University of Science and Technology\\
}

\maketitle

\begin{abstract}
	Recently, inspired by DETR variants, query-based end-to-end instance segmentation (QEIS) methods have outperformed CNN-based models on large-scale datasets. 
	Yet they would lose efficacy when only a small amount of training data is available since it's hard for the crucial queries/kernels to learn localization and shape priors.
	To this end, this work offers a novel unsupervised pre-training solution for low-data regimes.
    Inspired by the recent success of the Prompting technique, we introduce a new pre-training method that boosts QEIS models by giving \textbf{Saliency Prompt} for queries/kernels.
    Our method contains three parts: 1) \textbf{Saliency Masks Proposal} is responsible for generating pseudo masks from unlabeled images based on the saliency mechanism. 
    2) \textbf{Prompt-Kernel Matching} transfers pseudo masks into prompts and injects the corresponding localization and shape priors to the best-matched kernels.
	3) \textbf{Kernel Supervision} is applied to supply supervision at the kernel level for robust learning. 
   From a practical perspective, our pre-training method helps QEIS models achieve a similar convergence speed and comparable performance with CNN-based models in low-data regimes.
   Experimental results show that our method significantly boosts several QEIS models on three datasets.
   Code will be made available.
\end{abstract}

\section{Introduction}
\label{sec:intro}


\begin{figure}[t]
  \centering
  \includegraphics[width=1\linewidth]{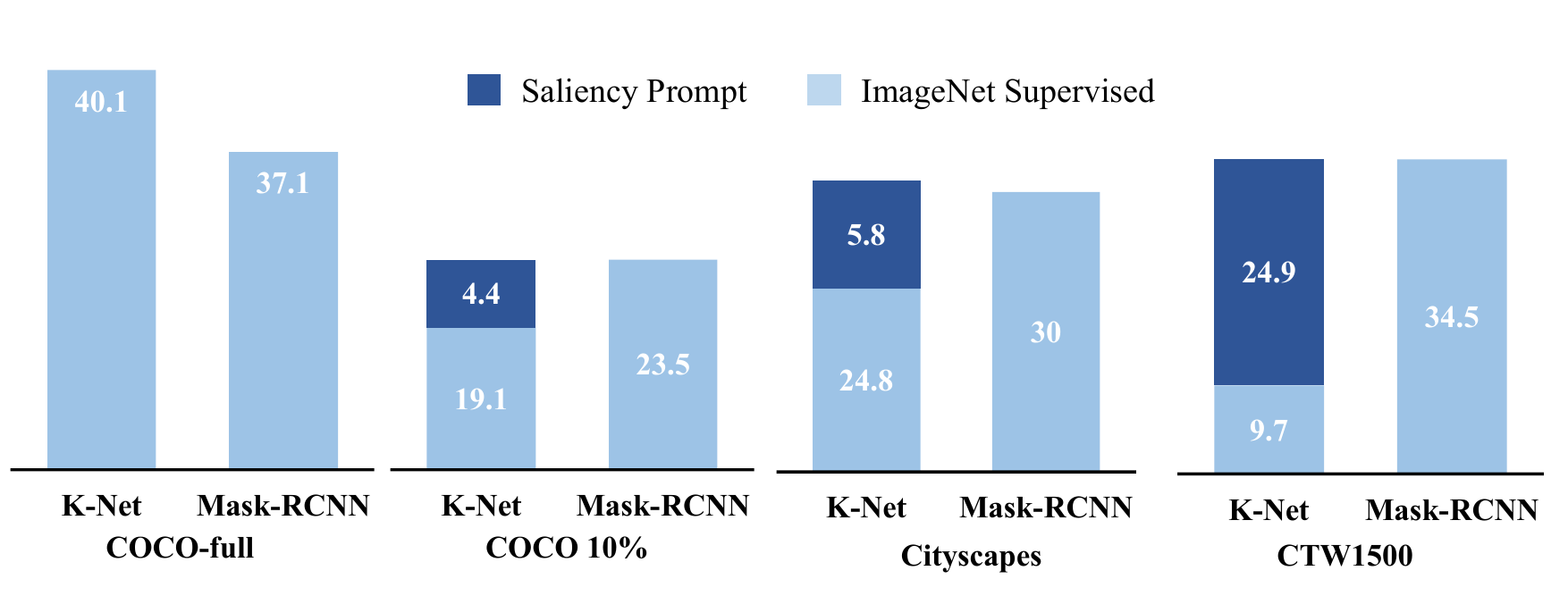}
  \caption{Performance comparison between K-Net and Mask-RCNN. K-Net can outperform Mask-RCNN on large-scale datasets (COCO-full). However, on small datasets (the right three), it can not perform as well as Mask-RCNN since it's hard to learn localization and shape priors. Our proposed unsupervised pre-training method based on saliency prompt not only boosts the vanilla K-Net significantly, but also helps to achieve comparable performance compared with Mask-RCNN.}
  \label{fig:tease}
\end{figure}

Modern CNN models address the instance segmentation task in an indirect way, by defining the localization problem on a large set of proposals\cite{maskrcnn}, window centers\cite{tensormask, instanceFCN}, or location-based masks\cite{fcos,solov1,solov2}.
A typical example is Mask-RCNN\cite{maskrcnn}, which generates candidate bounding boxes using a well-designed region proposal network. 
Although this paradigm makes localization learning easily optimized,
it still relies on the manually-designed non-maximum suppression (NMS) as post-processing to remove duplicated predictions.

Based on a state-of-the-art object detection model, DETR \cite{liu2021survey}, a few Query-based End-to-end Instance Segmentation (QEIS) models \cite{queryinst,mask2former,maskformer,maxdeeplab,istr,knet} have been proposed to perform instance segmentation in a new way.
Unlike CNN-based methods which usually require a large set of proposals, QEIS models use dynamic queries/kernels to automatically encode object localization knowledge with different locations and object shapes.
This design effectively eliminates hand-crafted anchors and post-processing like NMS.
However, due to the intrinsic dynamic attribute, the kernels are forced to learn general object spatial distribution and shape priors in a data-driven manner so that they can fit any input image.
This makes QEIS models require a much larger amount of training data and a much longer training time to achieve competitive performance with CNN-based methods.
Once in low-data regimes\cite{detreg}, QEIS models will encounter much more significant performance drops than CNN-based methods, as shown in Figure~\ref{fig:tease}. Here we take K-Net \cite{knet} as the typical example of QEIS models and compare it with Mask-RCNN.

That being said, the potential of QEIS models is still enormous since once good localization and shape priors can be learned, they can perform on par with or even outperform CNN-based methods with a much more concise pipeline. This makes us think about
how we can help QEIS models learn localization and shape priors quickly, especially for low-data regimes.

A promising solution is to adopt unsupervised pre-training, which requires no extra data and any modification to existing model architectures. However, most existing unsupervised pre-training methods \cite{detreg,updetr,swav,densecl,mocov2  } 
are only used for the backbone and can not benefit instance segmentation prediction heads, where localization and shape priors are exactly encoded.
In the object detection field, some works\cite{updetr,detreg} do pre-train a full detection architecture.
However, 
they use pseudo bounding boxes for training, many of which do not contain any object inside hence can not generate pseudo instance masks for instance segmentation.
FreeSOLO\cite{wang2022freesolo} is the first method specifically designed for instance segmentation. 
Yet it mainly focuses on generating pseudo masks and directly using them to supervise the model training. Such a way still learns the object localization and shape priors in a data-driven manner, hence requiring tedious steps to generate high-quality pseudo masks. 
	
To address these problems, we present a novel un-supervised pre-training method for QEIS models. Inspired by the recent advances in Prompting in NLP and vision tasks \cite{bert,clip,coop,cooop,vpt}, we propose to directly inject localization and shape priors into the kernels using our proposed \textbf{Saliency Prompt (SP)}.
The prompts are generated by saliency masks which indicate potential objects, and then are used to decorate the kernels for injecting location and shape knowledge.
    
In detail, our saliency prompt involves two essential parts: \textit{saliency} and \textit{prompt}:  
First, a \textbf{Saliency Mask Proposal} generation method is responsible for generating saliency-level pseudo masks from unlabeled images. 
Instead of directly learning from noisy pseudo masks, we use them to generate corresponding region features and then achieve prompts from them. 
Next, a \textbf{Prompt-Kernel Matching} module matches the saliency prompts to the kernels and then injects the prior knowledge encoded in the prompts into the best-matched kernels.
Furthermore, we also propose a \textbf{Kernel Supervision} scheme to supervise the model learning at the kernel level to gain kernel robustness. 
See Figure~\ref{fig:overview} for overview.

In our experiments, our method surpasses all the existing unsupervised pre-training algorithms on low-data regimes on four datasets.
It can be used as a plug-and-play pre-training step for most QEIS methods and
enables faster convergence speed and better performance without any increase in parameters or memory. 
Most importantly, our method achieves two \textit{desiderata} on downstream tasks
a) it leads to the same convergence speed as CNN-based methods.
(b) it gains comparable or even better performance than CNN-based methods on most downstream datasets.
In ablations, we find that our method shows big tolerance to the quality of pseudo masks. As such, we can easily achieve performance improvement without a sophisticated and time-consuming pseudo mask generation method as in FreeSOLO\cite{wang2022freesolo}.

\section{Related Work}
\subsection{Query-Based End-to-End Instance Segmentation}
    With the development of Transformer, a brand new object detector based on object-queries is proposed by DETR\cite{detr}. 
    It considers the detection task as a set prediction problem, which makes DETR become the first end-to-end model without any human-crafted anchors or NMS. 
    Subsequently, many works\cite{queryinst,mask2former,maskformer,maxdeeplab,istr,knet} follow this paradigm to tackle the instance segmentation task, which we call them Query-Based End-to-End Instance Segmentation (QEIS) methods. 
    In the prediction head, instead of proposing dense object proposals, QEIS models use queries/tokens/kernels to capture individual instance features on the global scale, hence are more flexible and enable end-to-end learning. Meanwhile, various improvements have been made by different QEIS models. 
    Inspired by the idea of SOLO-v2 \cite{solov2}, K-Net\cite{knet} generates convolution kernels to predict masks directly. 
    This kernel-mask paradigm enables K-Net to segment both semantic and instance categories consistently by a group of learnable kernels.
    QueryInst\cite{queryinst} builds upon Sparse-RCNN \cite{sparse} and adopts parallel supervision on dynamic mask heads.
    Mask2Former\cite{mask2former}  improves the efficiency and accuracy of the prediction head by using masked-cross-attention and multi-scale feature fusion.
	In this work, we take K-Net as a typical example of QEIS models and develop our unsupervised pre-training method upon it. However, our method can also be deployed on other QEIS-style models freely and improve their performance on low-data regimes, as proved in the experiments part.

\begin{figure*}[htbp]
  \centering
  \includegraphics[width=0.95\linewidth]{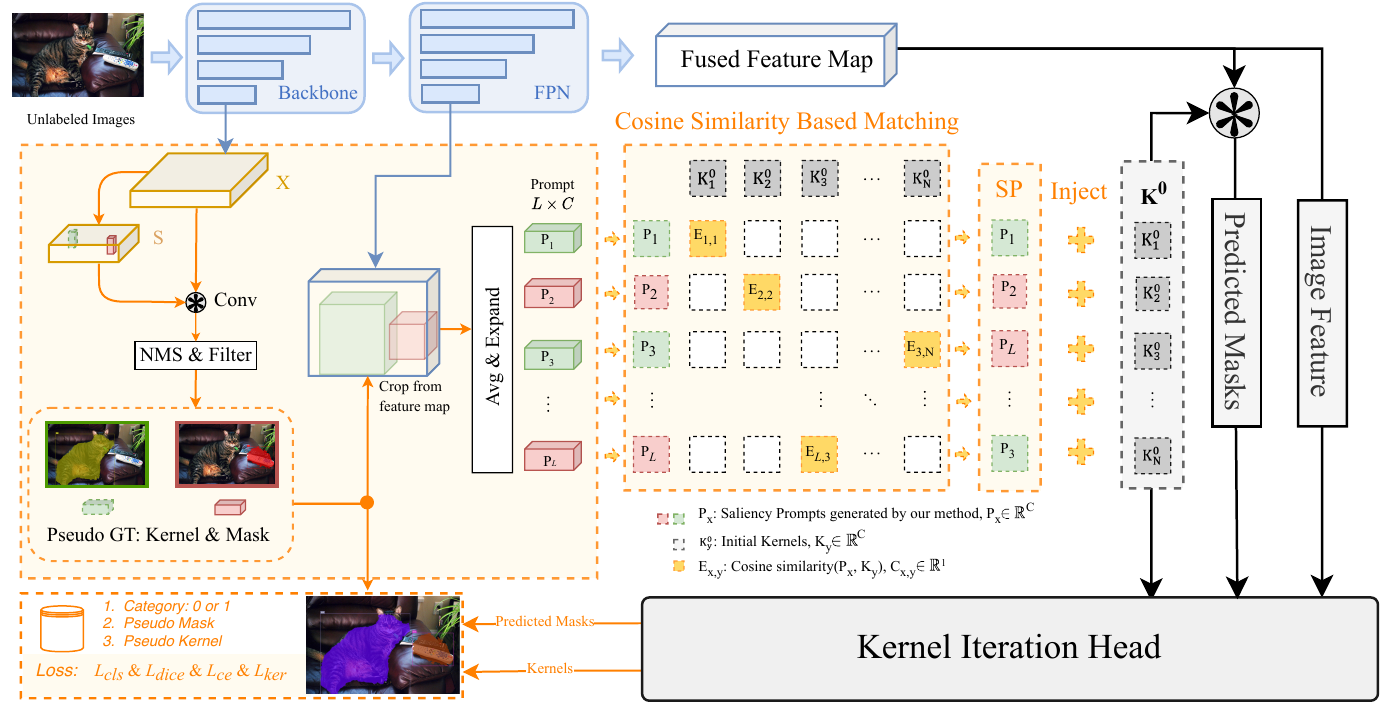}
  \caption{Overview of our proposed pre-training framework. 
  Modules with {\color[HTML]{F56B00} orange} colors denotes our pre-training method with the corresponding supervision. As can be seen,our method is parameter-free.
  {\color[HTML]{A9C4EB} Blue} and {\color[HTML]{CCCCCC} gray} modules denote a vanilla QEIS model, here we use K-Net for example. }
  \label{fig:overview}
    \vspace{-0.2in}
\end{figure*}


\subsection{Unsupervised Pre-training}
Unsupervised pre-training aims to pre-train deep models with carefully designed pretext tasks for boosting the model performance in downstream tasks.
Most state-of-the-art unsupervised pre-training methods, such as DenseCL\cite{densecl}, SwAV\cite{swav}, and MoCo-v2\cite{mocov2}, are only used to pre-train backbones and ignore the prediction head, thus can not solve the data-hungry problem of QEIS models. 

UP-DETR\cite{updetr} and DETReg\cite{detreg} creatively build end-to-end unsupervised learning frameworks based on DETR's query-object mechanism. 
Specifically, UP-DETR\cite{updetr} randomly crops image patches from images as pseudo labels. DETReg\cite{detreg} uses region proposals generated by ResNet and SwAV as pseudo labels.
However, none of these methods work for  segmentation tasks as their pseudo labels would only contain backgrounds that cannot provide object and shape information. Such pseudo labels can largely mislead the training of the segmentation models.

Recent work FreeSOLO\cite{wang2022freesolo} concentrates on generating good-quality pseudo masks inspired by SOLOv2 \cite{solov2}, making the first time, pseudo labels can be used for training instance segmentation models.
However, FreeSOLO requires multiple steps, such as pre-training and self-training, to generate pseudo masks. It only considers using the pseudo masks as labels to supervise the model training, without exploring how to use them in a more efficient way.
Our experiments indicate that explicitly injecting object localization and shape priors contained in the pseudo labels can bring further improvements compared with solely using pseudo labels as the supervision.

\subsection{Prompting}
	The prompting technique originates from NLP\cite{bert} and is soon transferred to the multi-modal domain\cite{clip,coop,cooop}. 
	It formulates downstream tasks as a "fill-in-the-bank" problem, such as "A photo of a \{object\}" in CLIP\cite{clip}.
	Here "A photo of a" stands for prompt templates, which guide the language model to elicit useful information from the pretrained models and predict the "\{object\}".
    CoOp\cite{coop} replaces man-defined language prompts with a set of learnable vectors. 
    Based on CoOp, CoCoOp\cite{cooop} is further purposed to generate input-conditional prompts for each image and combine them with existing language dynamic prompts.
	To conclude, prompting has shown its performance in the language and vision-language domain.
    Most recently,
    VPT \cite{vpt} first integrates prompting into pure vision tasks. 
    It prepends several learnable prompt tokens to the patch tokens of the frozen ViT\cite{vit} model to fit it for a variety of downstream tasks and datasets without the need of finetuning the whole ViT model.
    We find that most previous works use prompts to improve the performance of pre-trained models on downstream tasks. That is to say, they do not use prompts in the pre-training stage and only utilize them in the finetuning stage. Contrary to them, we ingeniously tailor the prompting mechanism for our pre-training task.
    Our saliency prompts are only used in the pre-training stage to help the QEIS model learn a better prediction head and then are removed in the finetuning stage.

\section{Methodology}
\label{sec:method}

In this section, we take K-Net \cite{knet} as a typical example of QEIS models and present our proposed unsupervised pre-training method upon it.  We first briefly review the K-Net model, and then show how to use our proposed saliency prompt for pre-training K-Net.

\subsection{K-Net Review}
\label{subsec:knet}
In most instance segmentation scenarios, the number of instances to segment is usually assumed to be unknown (average 7.7 in COCO). 
	K-Net dynamically encodes the instance-level information into $N$ kernels\footnote{$N$ is defined to be larger than the maximum instance number in images.}, each of which is responsible for ﬁnding the pixels belonging to its corresponding instance. 
	In particular, given the feature maps $\mathbf{F} \in \mathbb{R}^{C \times H \times W}$ of the input image and the generated kernels $\mathbf{K} \in \mathbb{R}^{N \times C}$, the instance segmentation masks $\mathbf{M} \in \mathbb{R}^{N \times H \times W}$ can be obtained by performing convolution on $\mathbf{F}$ with $\mathbf{K}$, denoted as
 \begin{equation}
 \mathbf{M}=\sigma(\mathbf{K} * \mathbf{F}),
 \end{equation}
 where $\sigma$ means the sigmoid activation function and $*$ denotes the convolution operation.
 
For kernel generation, K-Net designs a dynamic kernel update mechanism that uses the segmented masks and the features $\mathbf{F}$ to enhance the kernels $\mathbf{K}$ in an iterative way. At each iteration step $i$, the kernel updatation is formulated as
\begin{equation}
\mathbf{K}^i, \mathbf{M}^i=\mathcal{F}^i(\mathbf{M}^{i-1}, \mathbf{K}^{i-1}, \mathbf{F}).
\end{equation}
The initial kernels $\mathbf{K}^0$ are randomly initialized and then learned during training, encoding general image-agnostic localization and shape priors. The kernels are then updated by $\mathcal{F}^i$ to receive image-specific information.
Therefore, the learning of $\mathbf{K}^0$ can only be driven by the final instance segmentation loss and the localization loss, hence needs lots of training data and time to learn general priors.



\subsection{Saliency Mask Proposal Generation}
Most previous deep unsupervised segmentation methods generate pseudo masks with unsupervised algorithms for supervising the model training. Here we also follow the same pipeline and additionally use the pseudo masks for generating prompts.
There exist various unsupervised algorithms for generating pseudo masks, such as selective search \cite{selective}, random proposal\cite{updetr}, and FreeSOLO\cite{wang2022freesolo}. Here we adopt the saliency mechanism \cite{liu2020picanet,liu2021visual,zhuge2022salient,fang2021densely} since its simplicity.
Our main idea is to generate dense saliency maps through foreground-background separation modeling.
We first use a self-supervised pre-trained model, such as ResNet-50 \cite{resnet} trained with the DenseCL\cite{densecl} algorithm, as the backbone to extract image features.
Then, dense feature similarity is calculated upon the output of the backbone network to generate dense saliency maps.
Specifically, given the feature maps $\mathbf{X} \in \mathbb{R}^{H \times W \times D}$, 
we first uniformly sample $H' \times W'$ foreground seeds and generate the seed features $\mathbf{S} \in \mathbb{R}^{H' \times W' \times D}$ using average pooling on $\mathbf{X}$.
Next, dense saliency is computed by using the feature of each seed 
$\mathbf{S}_{i,j} \in \mathbb{R}^D$ as the convolution weights to convolve the feature $\mathbf{X}$. 
The operation for generating a saliency map $\mathbf{Y}_{i, j}$ for the seed $(i,j)$ is formulated as
\begin{equation}
    \mathbf{Y}_{i, j}=\operatorname{Conv}\left(\mathbf{S}_{i,j}. \mathbf{X}\right)\in \mathbb{R}^{H \times W}.
\end{equation}
The convolution operation calculates the similarity between the weights and the feature at each location. The locations obtain large convolution activation means they are similar to the foreground seed $(i,j)$ hence belonging to the salient foreground. While those that have small convolution activation belong to the background of the seed. Then,
we linearly normalize the saliency map $\mathbf{Y}_{i, j}$ to the range [0,1] and separate the foreground and background by using a threshold to binarize it, thus obtaining the saliency mask for the seed $(i,j)$. Finally, each foreground seed in $\mathbf{S}$ has a saliency mask, which usually highlights the coarse region of the foreground object this seed belongs to.\footnote{If one seed does not belong to any object, usually we get a null saliency mask.} However, different saliency masks may indicate the same object since their seeds may all belong to this object.
Hence, we further use the mask NMS to filter out overlapping masks. The whole process can be formulated as
\begin{equation}
    \mathbf{Z}=\operatorname{NMS}\left(\operatorname{Thres} \left(\operatorname{Norm}\left(\mathbf{Y}\right)\right)\right),
\end{equation}
where \textbf{Z} denotes the ﬁnal saliency mask proposals and $\operatorname{Norm}$ means linear normalization. The process of generating saliency mask proposals is also shown in Figure \ref{fig:overview}.

\subsection{Prompt-Kernel Matching}

	Figure \ref{fig:overview} shows the details of our proposed Prompt-Kernel Matching, which has two key steps: Prompt Generation and Cosine Similarity Based Matching.
 
\paragraph{Prompt generation.} 
	Given the saliency mask proposals $\mathbf{Z}$, we use their tightest bounding boxes to crop the image feature maps output by an FPN \cite{fpn}. We denote the cropped features as $\mathbf{f} = \{\mathbf{f}_1, \mathbf{f}_2, \cdots, \mathbf{f}_L\}$, where $L$ is the number of the masks in $\mathbf{Z}$ ($L$ can vary from different images). 
    Then, we use average pooling to convert the features $\mathbf{f}$ into prompts:
    \begin{equation}
    \mathbf{P} = \text{Avg}(\mathbf{f})\in \mathbb{R}^{L \times C},
    \end{equation}
    where $C$ is also the channel number of the FPN feature and Avg means average pooling along the spatial dimension.
    
\paragraph{Cosine similarity based matching.}
	Each prompt encodes the localization and shape priors for an individual object and can be injected into one of the initial kernels $\mathbf{K}^0$ of K-Net.
 This raises an interesting question: which prompt should be injected into which kernel?
 A straightforward way is to randomly or sequentially assign the $L$ prompts to $N$ tokens. However, as found in DETR \cite{detr} and K-Net \cite{knet}, under the dynamic learning training scheme, different kernels/queries encode the localization and shape priors of different image regions and objects with different shapes, while each kernel mainly learns a specific pattern of similar object shapes and locations. As a result, using random or sequential assignments of the prompts may easily inject totally different object localization and shape information into the same token in different training samples, hence making the learning of the initial tokens very unstable.
    To this end, we propose a novel prompt-kernel matching scheme based on cosine similarity to assign the best-matched prompt to each token.
    Specifically,  given the $L$ prompts $\mathbf{P} = \{\mathbf{P}_l\in \mathbb{R}^{C}\}_{l=1}^L$ and the $N$ initial kernels $\mathbf{K}^0 = \{\mathbf{K}^0_n\in \mathbb{R}^{C}\}_{n=1}^N$, we compute the cosine similarity between them  to build the similarity matrix $\mathbf{E} \in \mathbb{R}^{N\times L}$:
    \begin{equation}
    \mathbf{E}_{n,l} = \frac{\mathbf{K}^0_n}{\|\mathbf{K}^0_n\|_2} \cdot \frac{\mathbf{P}_l}{\left\|\mathbf{P}_l\right\|_2}.
    \end{equation}
    Then, for each kernel $n$, we select the best-matched prompt index $\delta(n)$ with the largest similarity score:
    \begin{equation}
    \delta(n) = \mathop{\arg\max}\limits_{l\in [1,...,L]}\mathbf{E}_{n,l}.
    \end{equation}
    Next, the best-matched prompt $\mathbf{P}_{\delta(n)}$ is injected into the kernel $n$ via summation:
\begin{equation}
    \mathbf{K}^{0'}_n = \mathbf{K}^0_n + \mathbf{P}_{\delta(n)}.
\end{equation}
Finally, the decorated initial kernels $\mathbf{K}^{0'}$ are fed into the prediction head of K-Net. 
As such, each kernel can get the best-matched localization and shape awareness to ease its learning.

\subsection{Loss Function and Kernel Supervision}
\begin{table}[t]
\scriptsize
\caption{
Instance segmentation fine-tune results on COCO with 5\% and 10\% annotated images based on K-Net.  \vspace{-8pt}
}
\centering
\begin{tabular}{cccccccc}
\hline
                     & Pre-train                                 & mAP                                   & $AP_{50}$                                & $AP_{75}$                                & $AP_{S}$                                & $AP_{M}$                                 & $AP_{L}$                                 \\ \hline
                             & Img. Sup.                                   & 14.8                                  & 29.1                                  & 13.7                                  & 4.3                                  & 15.5                                  & 24.4                                  \\
                             & DenseCL                                     & 16.7                                  & 31.2                                  & 15.9                                  & 5.1                                  & 17.5                                  & 27.7                                  \\
                             & SwAV                                        & 15.7                                  & 30.3                                  & 14.7                                  & 4.6                                  & 25.9                                  & 16.6                                  \\
                             & MoCo-v2                                     & 17                                    & 32                                    & 16.2                                  & 5.3                                  & 18.3                                  & 27.1                                  \\
\multirow{-5}{*}{\rotatebox{90}{5\% images}}  & \cellcolor[HTML]{EFEFEF}\textbf{SP(ours)} & \cellcolor[HTML]{EFEFEF}\textbf{19.9} & \cellcolor[HTML]{EFEFEF}\textbf{35.7} & \cellcolor[HTML]{EFEFEF}\textbf{19.9} & \cellcolor[HTML]{EFEFEF}\textbf{6.0} & \cellcolor[HTML]{EFEFEF}\textbf{21.0} & \cellcolor[HTML]{EFEFEF}\textbf{32.6} \\ \hline
                             & Img. Sup.                                   & 19.1                                  & 35.7                                  & 18.2                                  & 6.7                                  & 20                                    & 31.6                                  \\
                             & DenseCL                                     & 20.3                                  & 36.4                                  & 20.3                                  & 6.6                                  & 21.8                                  & 33.6                                  \\
                             & SwAV                                        & 18.9                                  & 34.8                                  & 18.3                                  & 6.8                                  & 20.8                                  & 30.6                                  \\
                             & MoCo-v2                                     & 20.7                                  & 37.7                                  & 20.4                                  & 6.4                                  & 22.1                                  & 34.2                                  \\
\multirow{-5}{*}{\rotatebox{90}{10\% images}} & \cellcolor[HTML]{EFEFEF}\textbf{SP(ours)} & \cellcolor[HTML]{EFEFEF}\textbf{23.5} & \cellcolor[HTML]{EFEFEF}\textbf{41.4} & \cellcolor[HTML]{EFEFEF}\textbf{23.7} & \cellcolor[HTML]{EFEFEF}\textbf{7.9} & \cellcolor[HTML]{EFEFEF}\textbf{24.8} & \cellcolor[HTML]{EFEFEF}\textbf{38.6} \\ \hline

\end{tabular}
\vspace{-0.05in}
\label{tab:knet-coco}
\end{table}

     We use the saliency mask proposals $\mathbf{Z}$ as pseudo labels to perform bipartite matching with the $N$ predictions of the tokens and then use the set prediction loss to supervise the pre-training, which is the same as the original K-Net \cite{knet}.
	The overall loss function $\mathcal{L}_K$ of K-Net is assembled by three components:  the focal loss \cite{focal} $\mathcal{L}_{cls}$ for classification, the Dice loss $\mathcal{L}_{dice}$ and cross-entropy loss $\mathcal{L}_{ce}$ for segmentation, which only consider supervising the predictions. Since the predictions are mainly generated by the kernels, we argue that we can directly supervise the kernels as a supplementary loss. 
 
    Specifically, for each mask proposal $\mathbf{Z}_l$ where $l\in [1,...,L]$, we can find its corresponding saliency seed and feature $\mathbf{S}_{l} \in \mathbb{R}^D$, which encodes the representative object information of this mask. Then, we transform its channel number to $C$ for supervising the embedding of the token whose prediction is matched with the proposal $\mathbf{Z}_l$ after bipartite matching. We denote the index of the token matched with $\mathbf{Z}_l$ as $n_l$. Then, the kernel supervision loss can be formulated as
    \begin{equation}
    \mathcal{L}_{ker} = \sum_{l=0}^{L} \sum_i (1 - \operatorname{Cos}(\operatorname{Linear}(\mathbf{S}_{l}), \mathbf{K}^i_{n_l})),
    \end{equation}
    where the supervision is adopted for every K-Net kernel update iteration step $i$ and summed over all mask proposals. $\operatorname{Linear}$ means a linear transformation to reduce the channel number to $C$ and $\operatorname{Cos}$ denotes the cosine similarity.
    
 Our final loss can be defined as
    \begin{equation}
    \mathcal{L}_K = \lambda_{cls}\mathcal{L}_{c l s} + \lambda_{dice}\mathcal{L}_{dice} + \lambda_{ce}\mathcal{L}_{ce} +\lambda_{ker} \mathcal{L}_{ker}.
    \end{equation}
where  $\lambda_{(\cdot)}$ are corresponding loss weights. Since our pseudo labels are class-agnostic, we use binary classification (foreground \textit{v.s.} background) for $\mathcal{L}_{cls}$. 
  
\section{Experiments}
\subsection{Implementation Details}
\begin{figure}[t]
  \centering
  \includegraphics[width=0.9\linewidth]{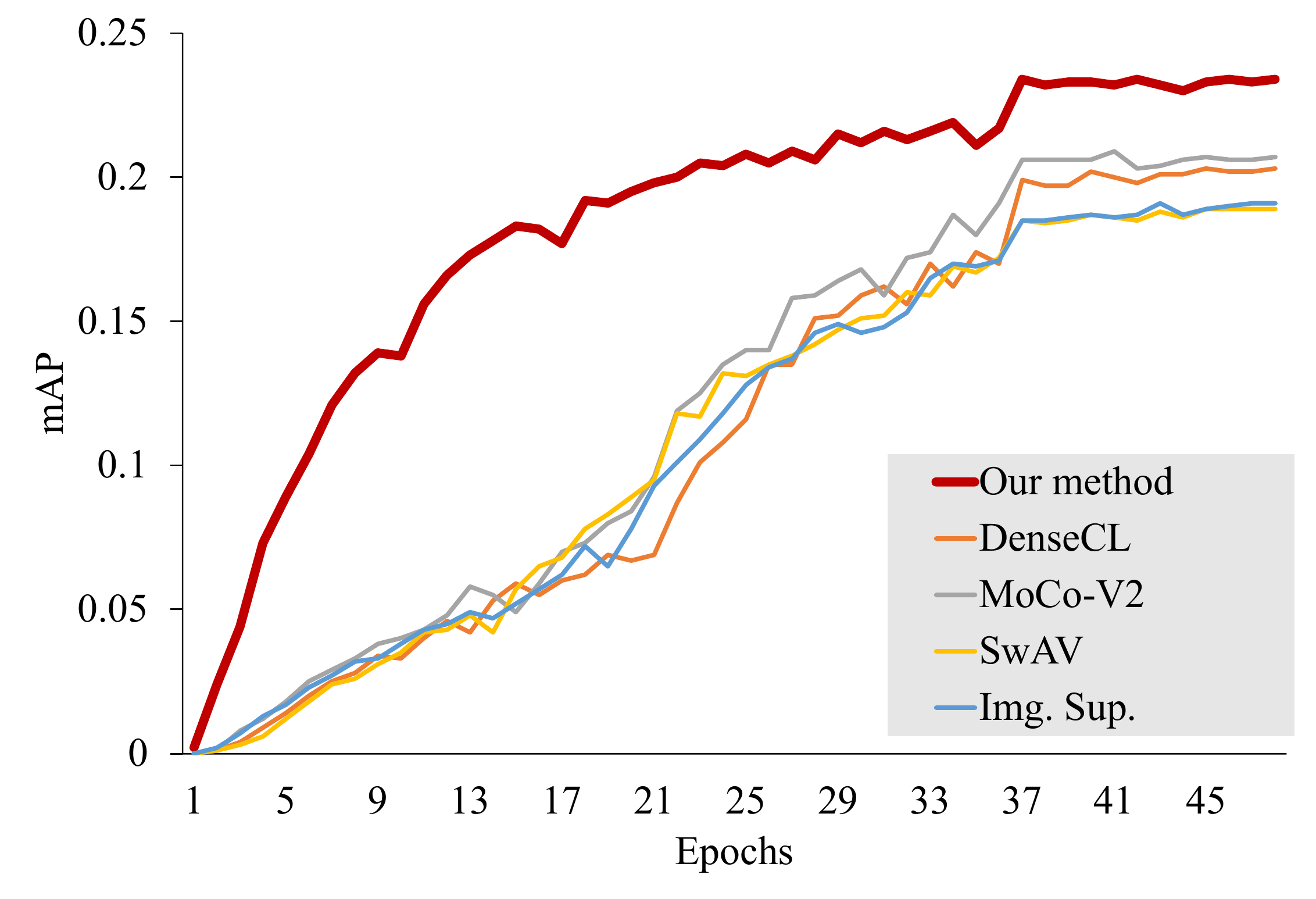}
  \vspace{-8pt}
  \caption{AP learning curves of K-Net with different pre-training methods on COCO with 10\% annotated images.}
  \label{fig:10image}
  \vspace{-8pt}
\end{figure}

\label{sec:exp}
\noindent \textbf{Pre-training setting.}
ResNet-50 \cite{resnet} is applied for all models as the backbone and pre-trained with the DenseCL algorithm.
We adopt the AdamW optimizer with $0.05$ weight decay and 1,000 steps of linear step warmup. 
As for data augmentation, we simply apply random flipping.
The model is trained with a batch size of 96 for 12 epochs on 8 A100 GPUs.
The initial learning rate is set to $1\times 10^{-4}$ and decreased by 0.1 after 8 and 11 epochs.
As for the hyperparameters of our model, we set the number of kernels/queries $N$ as 100,  $\mathcal{L}_{c l s}=2$,  $\mathcal{L}_{dice}=4$ , $\mathcal{L}_{ce}=1$ and  $\mathcal{L}_{ker}=1$.

\noindent \textbf{Fine-tuning setting.} 
All models are trained with a batch size of 96 on 8 A100 GPUs.
Random flipping and rotation are used as data augmentation.
Referring to the open-source of MMDetection\cite{mmdetection}, we use the same hyperparameters for QEIS and CNN-based models.
For QEIS, we apply the same training strategy as the pre-training stage except the training epoch, which will illustrate in the experiment tables. 
For CNN-based models, we apply SGD as the optimizer with weight decay and momentum. The learning rate is 0.02, momentum is set to 0.9 and weight decay is 0.0001.

\begin{table*}[t]
\scriptsize
\centering
\caption{Instance segmentation fine-tune results on Cityscapes and CTW1500.}\vspace{-8pt}
\label{tab:knet-other}
\begin{tabular}{cc|ccccccc|ccccccc}
\hline
                        &                                           & \multicolumn{7}{c|}{Cityscapes}                                                                                                                                                                                                                                    & \multicolumn{7}{c}{CTW1500}                                                                                                                                                                                                                                            \\
\multirow{-2}{*}{Model} & \multirow{-2}{*}{Pre-train}               & Epoch                               & AP                                    & $AP_{50}$                             & $AP_{75}$                          & $AP_{S}$                    & $AP_{M}$                            & $AP_{L}$                            & Epoch                               & AP                                    & $AP_{50}$                             & $AP_{75}$                           & $AP_{S}$                   & $AP_{M}$                              & $AP_{L}$                              \\ \hline
Mask RCNN\cite{maskrcnn}               & Img. Sup.                                      & 24                                  & 30                                    & 57.4                                  & -                                  & 8.3                         & 27.9                                & 49                                  & 96                                  & 34.5                                  & 69.8                                  & 32.1                                & 25.9                       & 40.4                                  & 36.8                                  \\
SOLO-v2\cite{solov2}                 & Img. Sup.                                      & 24                                  & 24.9                                  & 44.4                                  & -                                  & 1.8                         & 20.1                                & 50.4                                & 96                                  & 27.9                                  & 59.6                                  & 23.3                                & 8.7                        & 30.2                                  & 41                                    \\ \hline
                        & Img. Sup.                                      & 24                                  & 24.8                                  & 47.4                                  & -                                  & 4.8                         & 19.9                                & 43.1                                & 96                                  & 9.7                                   & 26.5                                  & 6.1                                 & 3.0                          & 9.2                                   & 19.2                                  \\
                        & DenseCL                                   & 24                                  & 28                                    & 52.2                                  & -                                  & 6.6                         & 25.2                                & 55.2                                & 96                                  & 18.9                                  & 42.6                                  & 15.1                                & 7.0                          & 18.8                                  & 32.5                                  \\
                        & SwAV                                      & 24                                  & 27.4                                  & 52.1                                  & -                                  & 5.3                         & 22.7                                & 49.2                                & 96                                  & 9.1                                   & 25.8                                  & 4.8                                 & 2.7                        & 9                                     & 19.8                                  \\
                        & MoCo-v2                                   & 24                                  & 28.2                                  & 51.2                                  & -                                  & 5.9                         & 26.4                                & 52.7                                & 96                                  & 13.3                                  & 32.2                                  & 10                                  & 4.3                        & 13.3                                  & 24.2                                  \\
\multirow{-5}{*}{K-Net} & \cellcolor[HTML]{EFEFEF}\textbf{SP(ours)} & \cellcolor[HTML]{EFEFEF}\textbf{24} & \cellcolor[HTML]{EFEFEF}\textbf{30.6} & \cellcolor[HTML]{EFEFEF}\textbf{55.4} & \cellcolor[HTML]{EFEFEF}\textbf{-} & \cellcolor[HTML]{EFEFEF}5.8 & \cellcolor[HTML]{EFEFEF}\textbf{27} & \cellcolor[HTML]{EFEFEF}\textbf{54} & \cellcolor[HTML]{EFEFEF}\textbf{96} & \cellcolor[HTML]{EFEFEF}\textbf{34.6} & \cellcolor[HTML]{EFEFEF}\textbf{71.1} & \cellcolor[HTML]{EFEFEF}\textbf{31} & \cellcolor[HTML]{EFEFEF}18.0 & \cellcolor[HTML]{EFEFEF}\textbf{36.1} & \cellcolor[HTML]{EFEFEF}\textbf{45.7} \\ \hline
\end{tabular}
\vspace{-0.05in}
\end{table*}
\begin{figure*}[t]
  \centering
  \begin{subfigure}{0.49\linewidth}
    \includegraphics[width=0.95\linewidth]{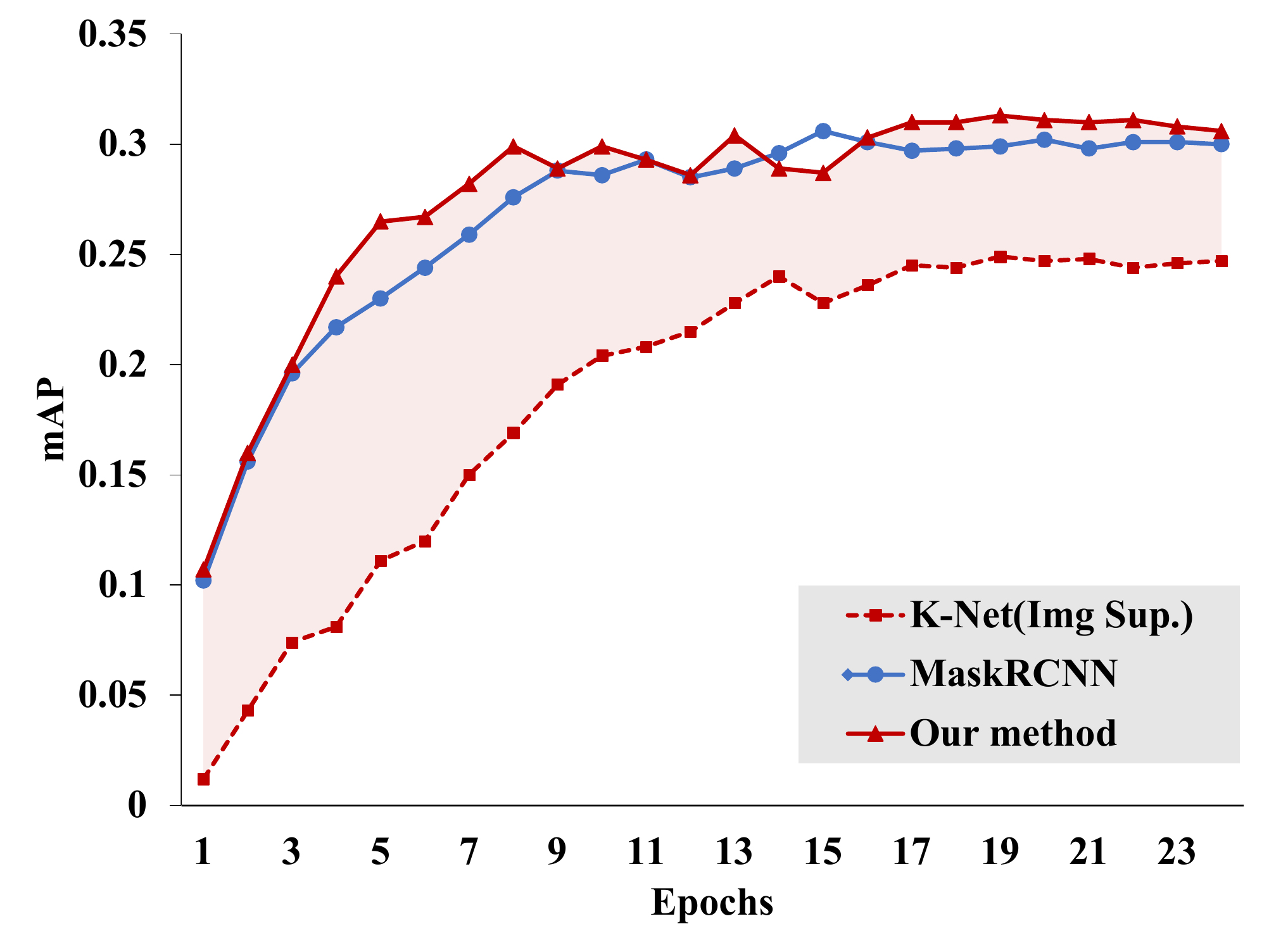}
    \caption{AP learning curves of Cityscapes}
  \end{subfigure}
  \begin{subfigure}{0.49\linewidth}
    \includegraphics[width=0.95\linewidth]{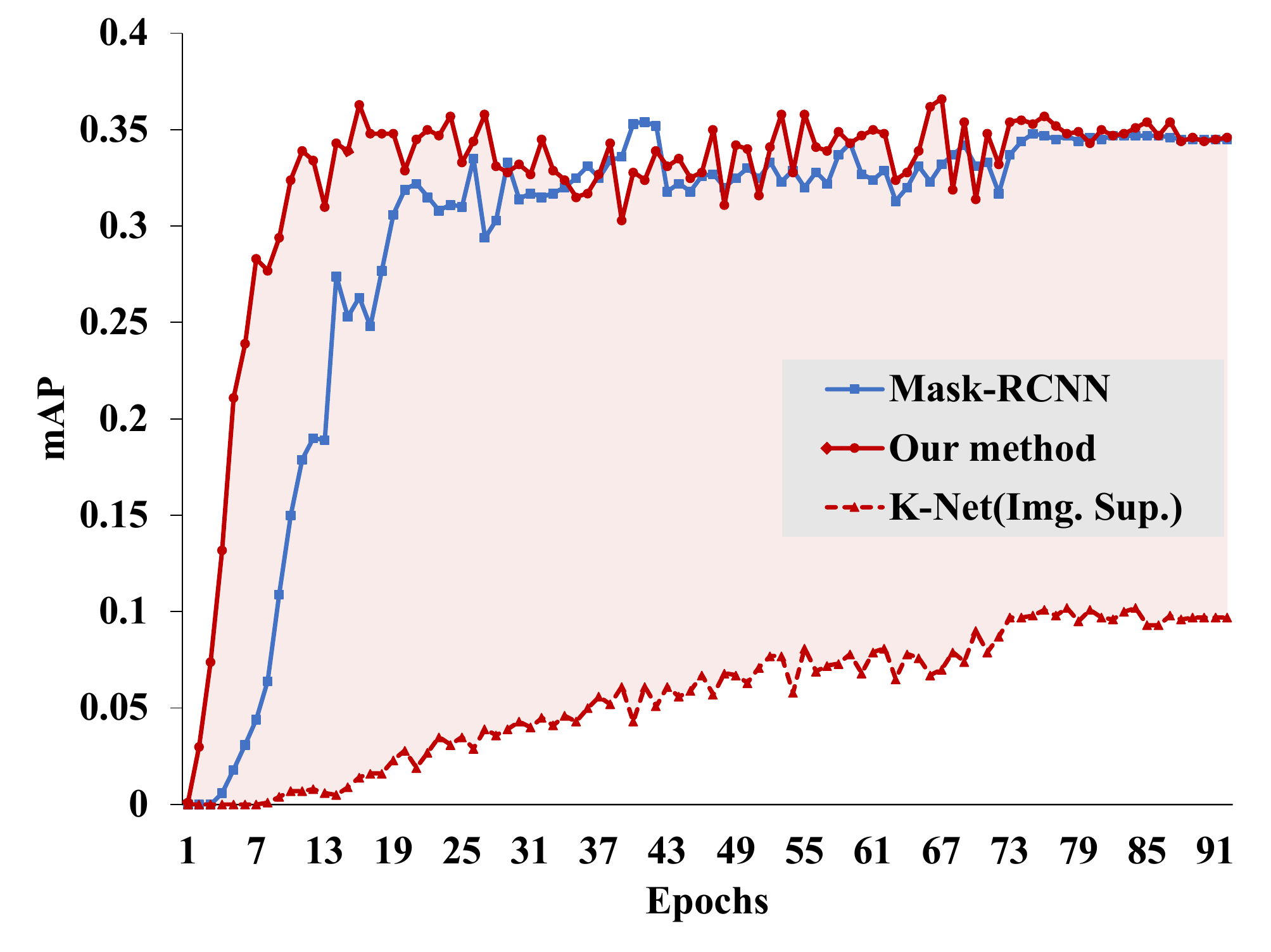}
    \caption{AP learning curves of CTW1500}
  \end{subfigure}
  \vspace{-4pt}
    \caption{AP learning curves of Mask-RCNN, vanilla K-Net and our method on Cityscapes and CTW1500.}
    \vspace{-10pt}
  \label{fig:curve-knet-other}
\end{figure*}

\noindent \textbf{Datasets.} We pretrain our QEIS models on MS COCO \cite{lin2014coco} unlabeled 2017 split and fine-tune on multiple datasets, including MS COCO, Cityscapes \cite{cityscapes} and CTW1500 \cite{ctw1500}.
{MS COCO} is a popular instance segmentation dataset that contains 164k labeled images, where objects from 80 object categories are annotated with dense pixel segmentation.
{Cityscapes} is a popular instance segmentation dataset which focus on semantic understanding and urban street scenes. It contains 5000 fine annotated large-scale images.
{CTW1500} is a wild-scene dataset that focuses on text detection and segmentation. It contains 1,500 images with dense annotation, which is also a typical low-data regime.

\subsection{Fine-tune Results on MS COCO}

To evaluate our performance in low-data regimes, we split the MS COCO train2017 dataset into two different types of training subset:

\begin{itemize}
    \item COCO with 10\% fully annotated images, which contains 12k+ images, 80k+ annotated masks.
    \item COCO with 5\% fully annotated images, which contains 5k+ images, 43k+ annotated masks.
\end{itemize}

Table \ref{tab:knet-coco} shows the comparison results on MS COCO with different pre-training methods. 
Img. Sup. denotes the ImageNet supervised pre-training.
As can be seen, vanilla K-Net performs poorly in low-data regimes. 
However, our pre-training method significantly boosts its performance compared with the ImageNet supervised pre-training: up to  \textbf{+5.1} AP on 5\% COCO and \textbf{+4.4} AP on 10\% COCO . 

Moreover, from the AP learning curves on 10\% COCO images shown in Figure \ref{fig:10image}, we can observe that our method converges much faster than other methods and gain much higher AP at the beginning of the fine-tuning stage.  
These evidences reflect that our method has probably already learned shape and localization prior in the pre-training stage.

\subsection{Fine-tune Results on Other Datasets}

In this part, we test our method in wild scenes with unseen targets (CTW1500\cite{ctw1500}) and small objects (Cityscapes\cite{cityscapes}). The comparison results are shown in
Table \ref{tab:knet-other}. 
As can be seen, 
our method outperforms the ImageNet-supervised method by \textbf{+5.8}AP and
achieves better performance(\textbf{+0.6}AP) compared with Mask-RCNN, a representative CNN-based model, on CItyscapes.
On CTW1500, our method surpasses the supervised and unsupervised methods by \textbf{+24.9}AP and \textbf{+15.7}AP. It also achieves comparable performance(\textbf{+0.1}AP) with Mask-RCNN.

These experiments further demonstrate that our method could enable K-Net to learn localization and shape prior rather than simply remembering objects during the pre-train stage. When compared with Mask-RCNN in small object scenarios in terms of $AP_{S}$, our method did not show superior results. We conjecture there are two reasons: 1) the intrinsic deficiency of the QEIS paradigm---both vanilla SOLOv2 and K-Net perform extremely badly compared with Mask R-CNN. 2) Saliency Mask Proposals mainly provide large-scale pseudo labels, which makes our kernels pay more attention to big objects rather than small objects.

Figure \ref{fig:curve-knet-other} shows the AP learning curves on Cityscapes and CTW1500. 
Although QEIS models like K-Net perform much worse than traditional CNN models like Mask-RCNN in both convergency speed and final accuracy,
equipped with our method, K-Net converges much faster than the ImageNet supervised pre-training models and can gain considerable learning curves compared with Mask-RCNN. 

\subsection{Deployed on QueryInst and Mask2Former}

\begin{table*}[t]
\caption{
Instance segmentation fine-tune results on QueryInst and Mask2Former.
}\vspace{-8pt}
\scriptsize
\centering
\begin{tabular}{cc|ccccccc|ccccccc}
\hline
                              &                                             & \multicolumn{7}{c}{CTW1500}                                                                                                                                                                                                                                                         & \multicolumn{7}{c}{Cityscapes}                                                                                                                                                                                                                                                 \\
\multirow{-2}{*}{Model}       & \multirow{-2}{*}{Pre-train}                 & Epoch                               & AP                                    & $AP_{50}$                             & $AP_{75}$                             & $AP_{S}$                              & $AP_{M}$                              & $AP_{L}$                              & Epoch                               & AP                                    & $AP_{50}$                             & $AP_{75}$                          & $AP_{S}$                             & $AP_{M}$                              & $AP_{L}$                             \\ \hline
                              & Img. Sup.                                        & 80                                  & 38.8                                  & 67.6                                  & 41.6                                  & 15.6                                  & 41.2                                  & 57.4                                  & 24                                  & 29.1                                  & 52.4                                  & -                                  & 5.6                                  & 23.9                                  & 55                                   \\
                              & DenseCL                                     & 80                                  & 43.2                                  & 71.6                                  & 48.5                                  & 18.4                                  & 47.6                                  & 59.9                                  & 24                                  & 27.5                                  & 48.9                                  & -                                  & 5.2                                  & 23.8                                  & 53.7                                 \\
                              & SwAV                                        & 80                                  &   41.2                            &      69.1                       &       46.1                       &      17.6                          &       45.2                          &        58.1                        & 24                                  & 30.3                                  & 53.3                                  & -                                  & 5.4                                  & 23.3                                  & 59                                   \\
                              & MoCo-v2                                     & 80                                  & 43.3                                  & 71.2                                  & 49.2                                  & 18.9                                  & 47.6                                  & 59.4                                  & 24                                  & 30.7                                  & 54.3                                  & -                                  & 5.4                                  & 25.5                                  & 56.4                                 \\
\multirow{-5}{*}{Mask2Former\cite{mask2former}} & \cellcolor[HTML]{EFEFEF}\textbf{SP(ours)} & \cellcolor[HTML]{EFEFEF}\textbf{20} & \cellcolor[HTML]{EFEFEF}\textbf{52.9} & \cellcolor[HTML]{EFEFEF}\textbf{83.4} & \cellcolor[HTML]{EFEFEF}\textbf{62.1} & \cellcolor[HTML]{EFEFEF}\textbf{29.4} & \cellcolor[HTML]{EFEFEF}\textbf{56.4} & \cellcolor[HTML]{EFEFEF}\textbf{67.6} & \cellcolor[HTML]{EFEFEF}\textbf{24} & \cellcolor[HTML]{EFEFEF}\textbf{31.8} & \cellcolor[HTML]{EFEFEF}\textbf{55.8} & \cellcolor[HTML]{EFEFEF}\textbf{-} & \cellcolor[HTML]{EFEFEF}5.1 & \cellcolor[HTML]{EFEFEF}\textbf{26.5} & \cellcolor[HTML]{EFEFEF}\textbf{59.0} \\ \hline
                              & Img. Sup.                                        & 80                                  & 28.3                                  & 53.7                                  & 28.6                                  & 9.8                                   & 29                                    & 41.8                                  & 24                                  & 29.1                                  & 53.2                                  & -                                  & 6.7                                  & 27.4                                  & 50.7                                 \\
                              & DenseCL                                     & 80                                  & 31.6                                  & 56.7                                  & 33.4                                  & 10.4                                  & 32.5                                  & 46.6                                  & 24                                  & 30.8                                  & 54.7                                  & -                                  & 8.6                                  & 28.9                                  & 54.5                                 \\
                              & SwAV                                        & 80                                  & 24.6                                  & 50                                    & 23.1                                  & 8.1                                   & 25                                    & 36.3                                  & 24                                  & 30.7                                  & 54.4                                  & -                                  & 7.9                                  & 28.5                                  & 53.9                                 \\
                              & MoCo-v2                                     & 80                                  & 31.6                                  & 56.8                                  & 32.8                                  & 12.6                                  & 32                                    & 45.8                                  & 24                                  & 31.4                                  & 54.4                                  & -                                  & 8.1                                  & 28.4                                  & 56.1                                 \\
\multirow{-5}{*}{QueryInst\cite{queryinst}}   & \cellcolor[HTML]{EFEFEF}\textbf{SP(ours)} & \cellcolor[HTML]{EFEFEF}\textbf{20} & \cellcolor[HTML]{EFEFEF}\textbf{39.2} & \cellcolor[HTML]{EFEFEF}\textbf{66.8} & \cellcolor[HTML]{EFEFEF}\textbf{43.1} & \cellcolor[HTML]{EFEFEF}\textbf{16.7} & \cellcolor[HTML]{EFEFEF}\textbf{42.2} & \cellcolor[HTML]{EFEFEF}\textbf{51.9} & \cellcolor[HTML]{EFEFEF}\textbf{24} & \cellcolor[HTML]{EFEFEF}\textbf{32.8} & \cellcolor[HTML]{EFEFEF}\textbf{57.3} & \cellcolor[HTML]{EFEFEF}\textbf{-} & \cellcolor[HTML]{EFEFEF}\textbf{8.8} & \cellcolor[HTML]{EFEFEF}\textbf{29.2} & \cellcolor[HTML]{EFEFEF}\textbf{57.0}  \\ \hline
\end{tabular}

\vspace{-0.2in}
\label{tab:query-mask2former}
\end{table*}

\begin{table}[t]
\scriptsize
\caption{Ablation of {kernel supervised learning}.} \vspace{-8pt}
\begin{tabular}{cccccccc}
\hline
Model          & $L_{ker}?$           & mAP           & $AP_{50}$        & $AP_{75}$        & $AP_{S}$        & $AP_{M}$         & $AP_{L}$         \\ \hline
\rowcolor[HTML]{EFEFEF} 
\textbf{K-Net} & {\color[HTML]{009901}\textbf{\Checkmark}} & \textbf{23.5} & \textbf{41.4} & \textbf{23.7} & \textbf{7.9} & \textbf{24.8} & \textbf{38.6} \\
K-Net          & {\color[HTML]{CB0000}\XSolidBrush}        & 23.1          & 41.4          & 23.0          & 7.5          & 24.6          & 37.8          \\ \hline
\end{tabular}

\label{tab:loss_sem-ablation}
\vspace{-0.3in}
\end{table}


Besides K-Net, here we further apply our pre-train method on another two QEIS methods:QueryInst\cite{queryinst} and Mask2Former\cite{mask2former}. 
As shown in table \ref{tab:query-mask2former}, with our pre-train method, QueryInst outperforms the state-of-the-art unsupervised pre-training method by \textbf{+7.6} AP on CTW1500 dataset and achieves gains of \textbf{1.4} AP on Cityscapes dataset.
For Mask2Former, our method achieves significant gains of \textbf{+ 9.6} AP over the state-of-the-art unsupervised pre-training method on CTW1500. 
%
%
%
These results indicate that our pre-training method can help the kernels/queries of QEIS models to learn localization and shape prior effectively and help gain competitive performance improvement.
%
%

\subsection{Ablation Study}
We perform ablation analysis to understand the impact of each component of our pre-train method.
In general, we pre-train K-Net with our method on COCO unlabeled2017 for 12 epochs and then fine-tune on COCO train2017 with 10\% images for 48 epochs.


\noindent \textbf{Loss function.} Firstly, we investigate the contribution of the proposed kernel supervision loss. As shown in table \ref{tab:loss_sem-ablation}, this loss function yields clear improvement, indicating that the kernel supervision plays a complementary role to prediction supervision---using the noisy prediction supervision alone may lead to over-fitting.

\noindent \textbf{Cosine Similarity Based Matching.} Table \ref{tab:prompt-ablation} shows the evaluation results of several prompt approaches, including Random Assignment, Sequential Assignment and Cosine Similarity based matching. 
Random Assignment is called 'shuffle' in UP-DETR\cite{updetr}, which leads to a performance drop of 0.4AP compared to the method even without using prompt. 
Sequential Assignment simply expands the number of saliency prompt and attaches them to the initial kernels of K-Net, which achieves 1.7 AP improvement compared to Random Assignment. 
Then, Cosine Similarity further surpasses Sequential Assignment by {0.5} AP. 

\begin{table}[t]
\scriptsize
\caption{Ablation of different prompt approaches. '\XSolidBrush' means the model only pre-trained by pseudo labels without prompt.}\vspace{-8pt}
\begin{tabular}{ccccccc}
\hline
Prompt   Approch  & mAP  & $AP_{50}$ & $AP_{75}$ & $AP_{S}$ & $AP_{M}$ & $AP_{L}$ \\
\hline
\XSolidBrush      & 21.7 & 38.7   & 21.7   & 7.2   & 23.5  & 36.2  \\
Random Assign     & 21.3 & 38.3   & 21.2   & 7.6   & 22.5  & 35.5  \\
Seq. Assgin   & 23   & 40.6   & 23     & 8     & 24.3  & 37.6  \\
\rowcolor[HTML]{EFEFEF} 
\textbf{Cosine Similarity} & \textbf{23.5} & \textbf{41.4} & \textbf{23.7} & \textbf{7.9} & \textbf{24.8} & \textbf{38.6}  \\\hline
\end{tabular}

\label{tab:prompt-ablation}
\vspace{-0.1in}
\end{table}


\begin{figure}[b]
  \vspace{-0.3in}
  \centering
  \begin{subfigure}{0.39\linewidth}
    \includegraphics[width=1\linewidth]{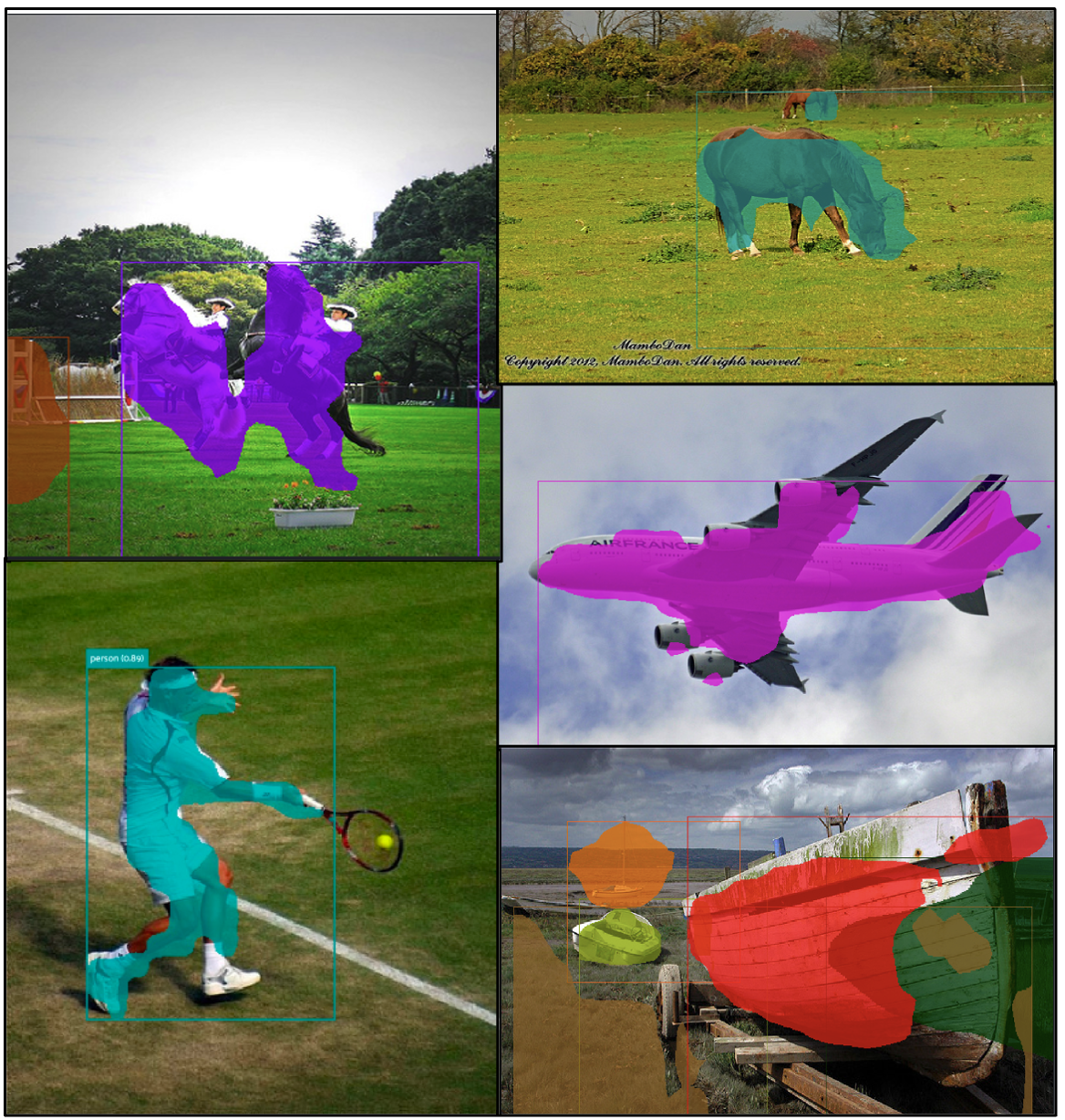}
    \caption{Saliency Masks Proposal.}
    \label{fig:pseudo-our}
  \end{subfigure}
  \hfill
  \begin{subfigure}{0.6\linewidth}
    \includegraphics[width=0.95\linewidth]{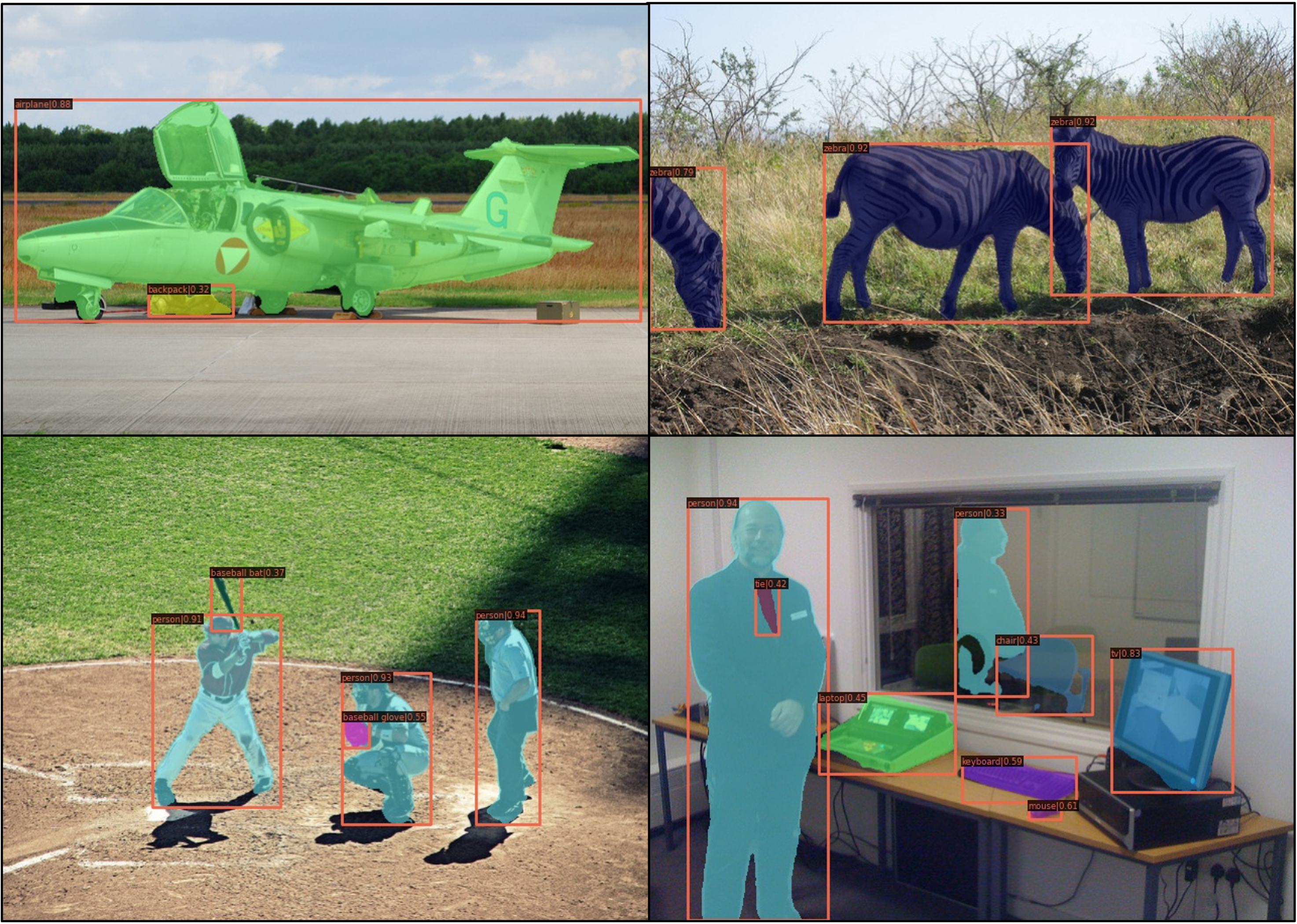}
    \caption{Fine-tune on COCO 10\% images.}
    \label{fig:pseudo-freesolo}
  \end{subfigure}
  \caption{Examples of our Saliency Mask Proposals and fine-tuned results.}
  \label{fig:psuedo-label}
  \vspace{-0.1in}
\end{figure}
\noindent \textbf{Class-agnostic Object Detection.}
We convert our proposed masks into bounding boxes like FreeSOLO\cite{wang2022freesolo}, and compare with UP-DETR\cite{updetr} and  DETReg\cite{detreg}.
Table \ref{tab:class_agnostic} shows the results of class-agnostic object detection on COCO val2017 benchmark. As can be seen, our method achieves better performance than other pre-training methods without self-training.
%
%
Although FreeSOLO has better performance than other methods, its self-training process requires much longer training time (extra 14 hours) and larger memory cost. 

\begin{table}[t]
\caption{Unsupervised class-agnostic object detection results.}\vspace{-8pt}
\centering
\footnotesize
\begin{tabular}{ccccc}
\hline
Method       & Self-train?                                & AP  & $AP_{50}$ & $AP_{75}$ \\ \hline
FreeSOLO\cite{wang2022freesolo}     & {\color[HTML]{009901} \Checkmark}          & 5.5 & 12.2    & 4.2     \\
UP-DETR\cite{updetr}      & {\color[HTML]{CB0000} \XSolidBrush}        & 0   & 0       & 0       \\
DETReg\cite{detreg}       & {\color[HTML]{CB0001} \XSolidBrush}        & 1.0   & 3.1     & 0.6     \\
\rowcolor[HTML]{EFEFEF} 
K-Net \textit{w} SP & {\color[HTML]{CB0002} \XSolidBrush}        & 3.2 & 8.5     & 2.0       \\ \hline
\end{tabular}

\label{tab:class_agnostic}
\vspace{-0.1in}
\end{table}

\noindent \textbf{Pseudo Mask Analysis.} 
Here we evaluate our Prompting method on three kinds of pseudo labels with different qualities. 
\begin{table}[t]
\centering
\caption{Ablation of Pseudo Labels on COCO with 10\% images. P means our Prompting method.} \vspace{-8pt}
\begin{tabular}{ccccc}
\hline
\rowcolor[HTML]{FFFFFF} 
\multicolumn{2}{l}{\cellcolor[HTML]{FFFFFF}Approch}                   & \multicolumn{2}{l}{\cellcolor[HTML]{FFFFFF}Pseudo Label (quality)} & mAP  \\ \hline
\multicolumn{2}{l}{K-Net $w/o$ P}                       & \multicolumn{2}{l}{Rand. Prop. (bad)}                               & 0.8 \\
\multicolumn{2}{l}{K-Net $w$ P}                         & \multicolumn{2}{l}{Rand. Prop. (bad)}                               & 10.0 \\ 
\multicolumn{2}{l}{K-Net $w/o$ P}                       & \multicolumn{2}{l}{Saliency (normal)}                              & 21.7 \\
\rowcolor[HTML]{EFEFEF} 
\multicolumn{2}{l}{\cellcolor[HTML]{EFEFEF}K-Net {\color[HTML]{009901}$w$ P}}    & \multicolumn{2}{l}{\cellcolor[HTML]{EFEFEF}{\color[HTML]{009901}Saliency (normal)}}      & 23.5 \\
\rowcolor[HTML]{EFEFEF} 
\multicolumn{2}{l}{\cellcolor[HTML]{EFEFEF}K-Net {\color[HTML]{CB0000}$w/o$ P}}  & \multicolumn{2}{l}{\cellcolor[HTML]{EFEFEF}{\color[HTML]{CB0000}FreeOLO (good)}}      & 23.3 \\
\multicolumn{2}{l}{K-Net $w$ P}                         & \multicolumn{2}{l}{FreeSOLO (good)}                               & 24.0 \\ \hline
\end{tabular}

\label{tab:pseudo-label}
\vspace{-0.25in}
\end{table}
As illustrated in table \ref{tab:pseudo-label}, pseudo labels generated by 'Random Proposal' have a negative impact on the results of the fine-tuned model, 
yet our pre-training method achieves significant improvement by utilizing Saliency Prompt.
Moreover, our Prompting method achieves competitive performance with both normal and good-quality pseudo masks, improving AP by 1.8 and 0.7, respectively. Besides, our method can also approach the performance of FreeSOLO which 
requires a much longer training time and many process steps. Notably, with the proposed Prompting approach, pre-train on normal pseudo masks gains comparable or even better performance than pre-train on good yet time-consuming pseudo masks like FreeSOLO. 
 We further visualize our pseudo masks in Figure \ref{fig:psuedo-label}.

\begin{figure*}[ht]
  \centering
  \begin{subfigure}{0.24\linewidth}
    \includegraphics[width=0.95\linewidth]{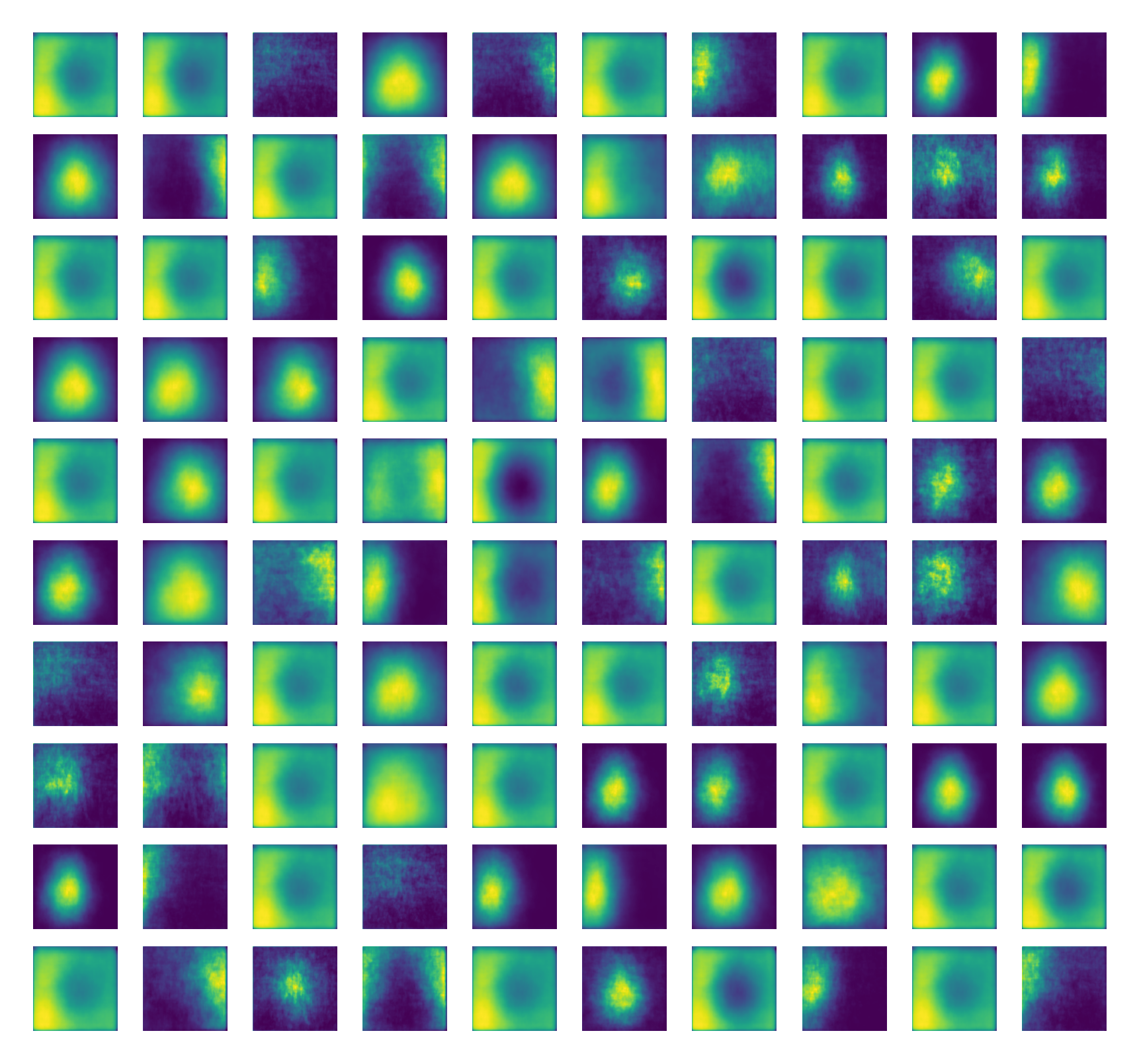}
    \caption{Train from scratch.}
    \label{fig:short-a}
  \end{subfigure}
  \hfill
  \begin{subfigure}{0.24\linewidth}
    \includegraphics[width=0.95\linewidth]{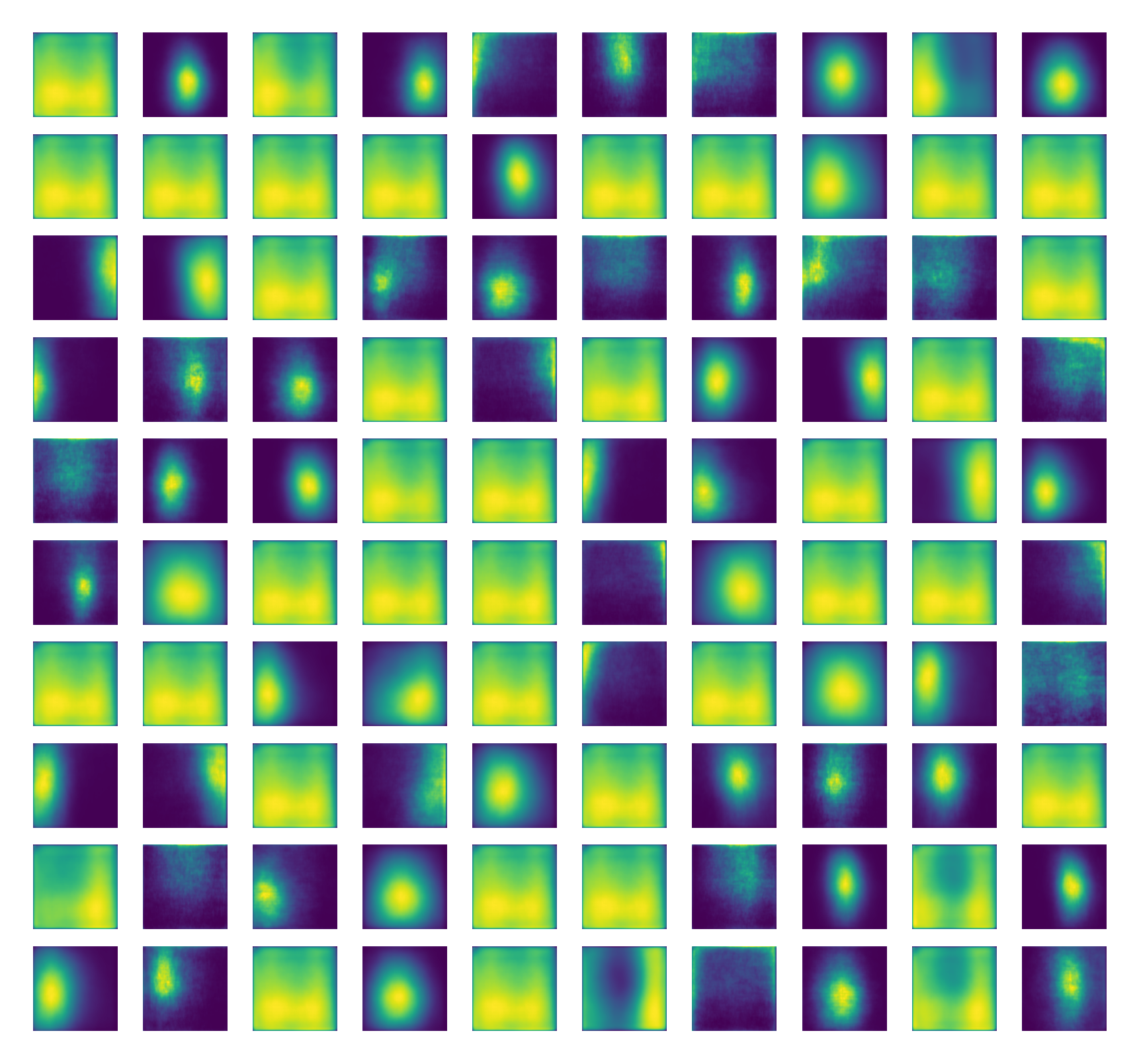}
    \caption{Pre-trained by MoCo-v2.}
    \label{fig:short-b}
  \end{subfigure}
    \begin{subfigure}{0.24\linewidth}
    \includegraphics[width=0.95\linewidth]{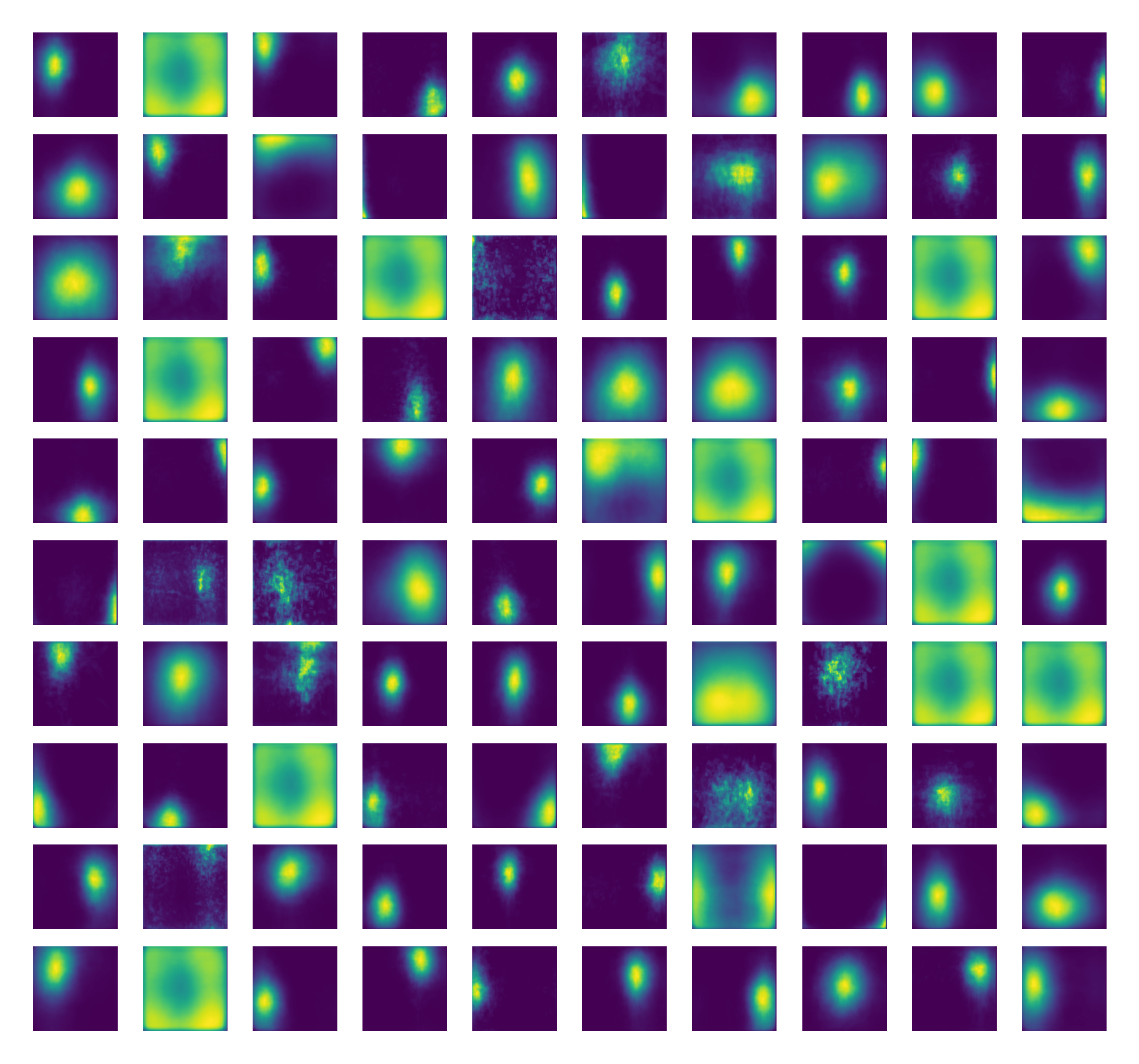}
    \caption{Pre-trained by our method.}
    \label{fig:short-c}
  \end{subfigure}
  \hfill
  \begin{subfigure}{0.24\linewidth}
    \includegraphics[width=0.95\linewidth]{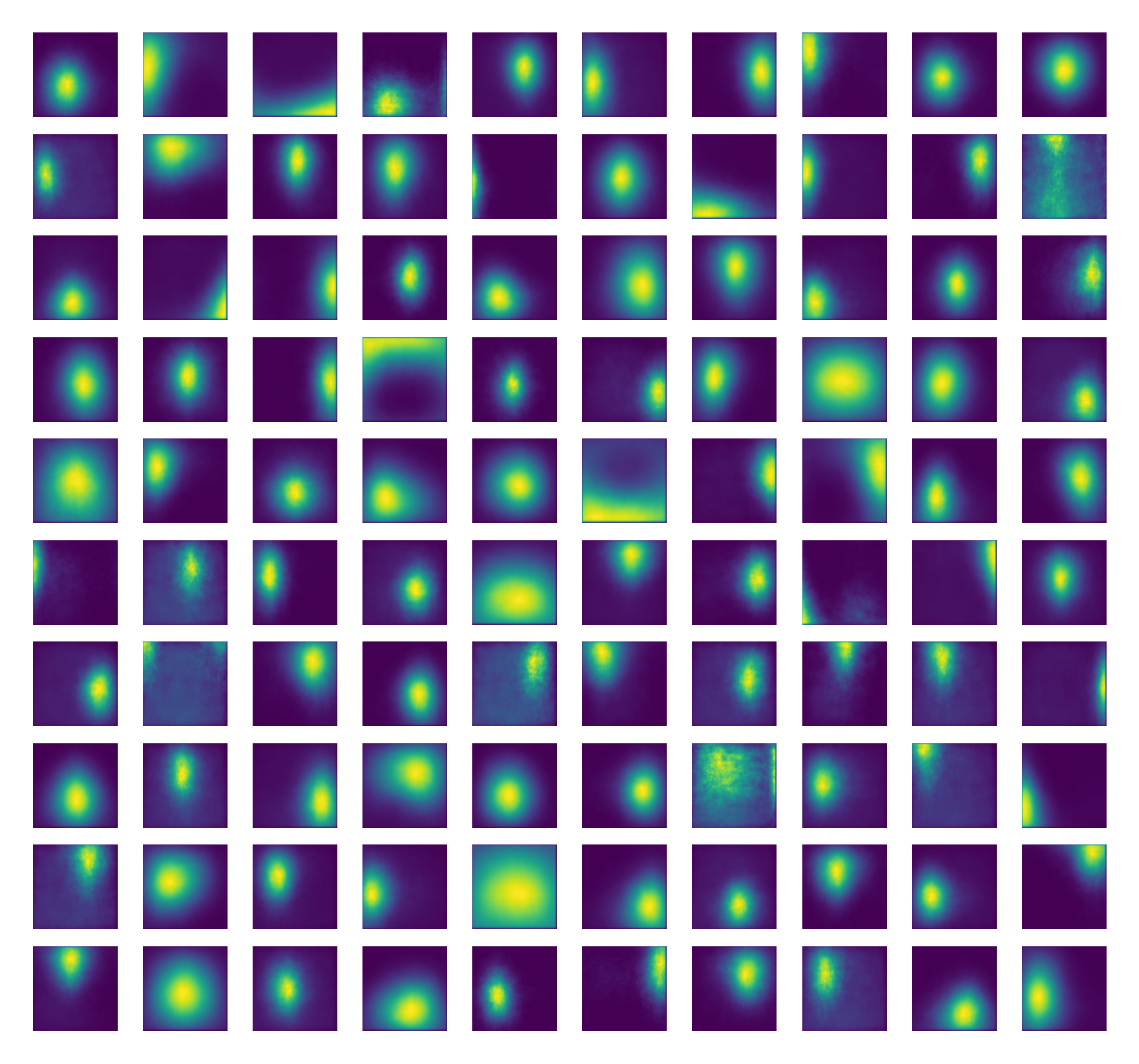}
    \caption{Trained on COCO train2017-full.}
    \label{fig:short-d}
  \end{subfigure} \vspace{-8pt}
  \caption{Average activation of the 100 kernels over 5000 images on COCO val2017. All masks are resized to $200\times 200$ for analysis.}\vspace{-8pt}
  \label{fig:visualize}
\end{figure*}

\subsection{Kernel Spatial Distribution Analysis}
	To further justify whether the proposed saliency prompt can assist the kernels of the prediction head to learn localization and shape priors, we visualize the average of mask activations of the 100 instance kernels over 5000 images on the 2017val split after 2 training epochs. 

	The best result (Figure~\ref{fig:short-d}) is from the fully trained K-Net on COCO train2017. 
	Those kernels have learned different shape and location priors. 
    However, the priors from the supervised (Figure~\ref{fig:short-a}) and the compared unsupervised (Figure~\ref{fig:short-b}) method mainly focus on the central area.
    Surprisingly, the kernels learned by our pre-train method (Figure~\ref{fig:short-c}) show positive trends in the diversity of shape and location priors, which is close to the fully trained kernels.
	These results demonstrate that the kernels pre-trained with Saliency Prompt have 
 learned effective spatial distribution and shape discrimination ability.

\section{Conclusion}

This paper first points out that the QEIS models lack spatial distribution and shape awareness and perform poorly in low-data regimes.
Hence we present \textbf{Saliency Prompt}, a novel unsupervised pre-train method using visual prompt, which can significantly boost the performance of QEIS models on low-data instance segmentation and achieve comparable or even better performance compared with CNN-based models.
From the perspective of technical, it is the first paper that explores the application of prompting in the instance segmentation field. We hope its novel design elements provide insights for future works on visual-based prompting mechanisms. In the future, we will follow more recent studies on visual saliency \cite{zhang2019synthesizing,liu2021scg,liu2021visual} for further promoting the prompt learning mechanism and apply the prompt learning mechanism to advance the weakly supervised learning community \cite{zhang2020weakly,zhang2021weakly,zhao2021weakly}. 

\noindent \textbf{Limitations.}
Most of our saliency masks are large-scale and simple textures, which makes our pre-trained kernels/queries mostly focus on large objects rather than small objects. 
Compared with the accuracy improvements on large objects, our pre-train method achieves limited improvement on small ones. 
We believe there is plenty of room to further optimize our proposed method.

{\small
\bibliographystyle{ieee_fullname}
\bibliography{egbib}

\begin{thebibliography}{10}\itemsep=-1pt

\bibitem{detreg}
Amir Bar, Xin Wang, Vadim Kantorov, Colorado~J Reed, Roei Herzig, Gal Chechik,
  Anna Rohrbach, Trevor Darrell, and Amir Globerson.
\newblock Detreg: Unsupervised pretraining with region priors for object
  detection.
\newblock In {\em Proceedings of the IEEE/CVF Conference on Computer Vision and
  Pattern Recognition}, pages 14605--14615, 2022.

\bibitem{detr}
Nicolas Carion, Francisco Massa, Gabriel Synnaeve, Nicolas Usunier, Alexander
  Kirillov, and Sergey Zagoruyko.
\newblock End-to-end object detection with transformers.
\newblock In {\em European conference on computer vision}, pages 213--229.
  Springer, 2020.

\bibitem{swav}
Mathilde Caron, Ishan Misra, Julien Mairal, Priya Goyal, Piotr Bojanowski, and
  Armand Joulin.
\newblock Unsupervised learning of visual features by contrasting cluster
  assignments.
\newblock {\em Advances in Neural Information Processing Systems},
  33:9912--9924, 2020.

\bibitem{mmdetection}
Kai Chen, Jiaqi Wang, Jiangmiao Pang, Yuhang Cao, Yu Xiong, Xiaoxiao Li,
  Shuyang Sun, Wansen Feng, Ziwei Liu, Jiarui Xu, Zheng Zhang, Dazhi Cheng,
  Chenchen Zhu, Tianheng Cheng, Qijie Zhao, Buyu Li, Xin Lu, Rui Zhu, Yue Wu,
  Jifeng Dai, Jingdong Wang, Jianping Shi, Wanli Ouyang, Chen~Change Loy, and
  Dahua Lin.
\newblock {MMDetection}: Open mmlab detection toolbox and benchmark.
\newblock {\em arXiv preprint arXiv:1906.07155}, 2019.

\bibitem{mocov2}
Xinlei Chen, Haoqi Fan, Ross Girshick, and Kaiming He.
\newblock Improved baselines with momentum contrastive learning.
\newblock {\em arXiv preprint arXiv:2003.04297}, 2020.

\bibitem{tensormask}
Xinlei Chen, Ross Girshick, Kaiming He, and Piotr Dollar.
\newblock Tensormask: A foundation for dense object segmentation.
\newblock In {\em 2019 IEEE/CVF International Conference on Computer Vision
  (ICCV)}, pages 2061--2069, 2019.

\bibitem{mask2former}
Bowen Cheng, Ishan Misra, Alexander~G Schwing, Alexander Kirillov, and Rohit
  Girdhar.
\newblock Masked-attention mask transformer for universal image segmentation.
\newblock In {\em Proceedings of the IEEE/CVF Conference on Computer Vision and
  Pattern Recognition}, pages 1290--1299, 2022.

\bibitem{maskformer}
Bowen Cheng, Alex Schwing, and Alexander Kirillov.
\newblock Per-pixel classification is not all you need for semantic
  segmentation.
\newblock {\em Advances in Neural Information Processing Systems},
  34:17864--17875, 2021.

\bibitem{cityscapes}
Marius Cordts, Mohamed Omran, Sebastian Ramos, Timo Rehfeld, Markus Enzweiler,
  Rodrigo Benenson, Uwe Franke, Stefan Roth, and Bernt Schiele.
\newblock The cityscapes dataset for semantic urban scene understanding.
\newblock In {\em Proc. of the IEEE Conference on Computer Vision and Pattern
  Recognition (CVPR)}, 2016.

\bibitem{instanceFCN}
Jifeng Dai, Kaiming He, Yi Li, Shaoqing Ren, and Jian Sun.
\newblock Instance-sensitive fully convolutional networks.
\newblock In Bastian Leibe, Jiri Matas, Nicu Sebe, and Max Welling, editors,
  {\em Computer Vision -- ECCV 2016}, pages 534--549, Cham, 2016. Springer
  International Publishing.

\bibitem{updetr}
Zhigang Dai, Bolun Cai, Yugeng Lin, and Junying Chen.
\newblock Up-detr: Unsupervised pre-training for object detection with
  transformers.
\newblock In {\em Proceedings of the IEEE/CVF conference on computer vision and
  pattern recognition}, pages 1601--1610, 2021.

\bibitem{bert}
Jacob Devlin, Ming-Wei Chang, Kenton Lee, and Kristina Toutanova.
\newblock Bert: Pre-training of deep bidirectional transformers for language
  understanding.
\newblock {\em arXiv preprint arXiv:1810.04805}, 2018.

\bibitem{vit}
Alexey Dosovitskiy, Lucas Beyer, Alexander Kolesnikov, Dirk Weissenborn,
  Xiaohua Zhai, Thomas Unterthiner, Mostafa Dehghani, Matthias Minderer, Georg
  Heigold, Sylvain Gelly, et~al.
\newblock An image is worth 16x16 words: Transformers for image recognition at
  scale.
\newblock {\em arXiv preprint arXiv:2010.11929}, 2020.

\bibitem{fang2021densely}
Chaowei Fang, Haibin Tian, Dingwen Zhang, Qiang Zhang, Jungong Han, and Junwei
  Han.
\newblock Densely nested top-down flows for salient object detection.
\newblock {\em arXiv preprint arXiv:2102.09133}, 2021.

\bibitem{queryinst}
Yuxin Fang, Shusheng Yang, Xinggang Wang, Yu Li, Chen Fang, Ying Shan, Bin
  Feng, and Wenyu Liu.
\newblock Instances as queries.
\newblock In {\em Proceedings of the IEEE/CVF International Conference on
  Computer Vision}, pages 6910--6919, 2021.

\bibitem{maskrcnn}
Kaiming He, Georgia Gkioxari, Piotr Doll{\'a}r, and Ross Girshick.
\newblock Mask r-cnn.
\newblock In {\em Proceedings of the IEEE international conference on computer
  vision}, pages 2961--2969, 2017.

\bibitem{resnet}
Kaiming He, Xiangyu Zhang, Shaoqing Ren, and Jian Sun.
\newblock Deep residual learning for image recognition.
\newblock In {\em Proceedings of the IEEE conference on computer vision and
  pattern recognition}, pages 770--778, 2016.

\bibitem{istr}
Jie Hu, Liujuan Cao, Yao Lu, ShengChuan Zhang, Yan Wang, Ke Li, Feiyue Huang,
  Ling Shao, and Rongrong Ji.
\newblock Istr: End-to-end instance segmentation with transformers.
\newblock {\em arXiv preprint arXiv:2105.00637}, 2021.

\bibitem{vpt}
Menglin Jia, Luming Tang, Bor-Chun Chen, Claire Cardie, Serge Belongie, Bharath
  Hariharan, and Ser-Nam Lim.
\newblock Visual prompt tuning.
\newblock {\em arXiv preprint arXiv:2203.12119}, 2022.

\bibitem{fpn}
Tsung-Yi Lin, Piotr Doll{\'a}r, Ross Girshick, Kaiming He, Bharath Hariharan,
  and Serge Belongie.
\newblock Feature pyramid networks for object detection.
\newblock In {\em Proceedings of the IEEE conference on computer vision and
  pattern recognition}, pages 2117--2125, 2017.

\bibitem{focal}
Tsung-Yi Lin, Priya Goyal, Ross Girshick, Kaiming He, and Piotr Doll{\'a}r.
\newblock Focal loss for dense object detection.
\newblock In {\em Proceedings of the IEEE international conference on computer
  vision}, pages 2980--2988, 2017.

\bibitem{lin2014coco}
Tsung-Yi Lin, Michael Maire, Serge Belongie, James Hays, Pietro Perona, Deva
  Ramanan, Piotr Doll{\'a}r, and C~Lawrence Zitnick.
\newblock Microsoft coco: Common objects in context.
\newblock In {\em ECCV}, pages 740--755. Springer, 2014.

\bibitem{liu2020picanet}
Nian Liu, Junwei Han, and Ming-Hsuan Yang.
\newblock Picanet: Pixel-wise contextual attention learning for accurate
  saliency detection.
\newblock {\em IEEE TIP}, 29:6438--6451, 2020.

\bibitem{liu2021visual}
Nian Liu, Ni Zhang, Kaiyuan Wan, Ling Shao, and Junwei Han.
\newblock Visual saliency transformer.
\newblock In {\em ICCV}, pages 4722--4732, 2021.

\bibitem{liu2021scg}
Nian Liu, Wangbo Zhao, Ling Shao, and Junwei Han.
\newblock Scg: Saliency and contour guided salient instance segmentation.
\newblock {\em IEEE Transactions on Image Processing}, 30:5862--5874, 2021.

\bibitem{ctw1500}
Yuliang Liu, Lianwen Jin, Shuaitao Zhang, and Sheng Zhang.
\newblock Detecting curve text in the wild: New dataset and new solution.
\newblock {\em CoRR}, abs/1712.02170, 2017.

\bibitem{liu2021survey}
Yang Liu, Yao Zhang, Yixin Wang, Feng Hou, Jin Yuan, Jiang Tian, Yang Zhang,
  Zhongchao Shi, Jianping Fan, and Zhiqiang He.
\newblock A survey of visual transformers.
\newblock {\em arXiv preprint arXiv:2111.06091}, 2021.

\bibitem{clip}
Alec Radford, Jong~Wook Kim, Chris Hallacy, Aditya Ramesh, Gabriel Goh,
  Sandhini Agarwal, Girish Sastry, Amanda Askell, Pamela Mishkin, Jack Clark,
  et~al.
\newblock Learning transferable visual models from natural language
  supervision.
\newblock In {\em International Conference on Machine Learning}, pages
  8748--8763. PMLR, 2021.

\bibitem{sparse}
Peize Sun, Rufeng Zhang, Yi Jiang, Tao Kong, Chenfeng Xu, Wei Zhan, Masayoshi
  Tomizuka, Lei Li, Zehuan Yuan, Changhu Wang, et~al.
\newblock Sparse r-cnn: End-to-end object detection with learnable proposals.
\newblock In {\em Proceedings of the IEEE/CVF conference on computer vision and
  pattern recognition}, pages 14454--14463, 2021.

\bibitem{fcos}
Zhi Tian, Chunhua Shen, Hao Chen, and Tong He.
\newblock Fcos: Fully convolutional one-stage object detection.
\newblock {\em arXiv preprint arXiv:1904.01355}, 2019.

\bibitem{selective}
Jasper~RR Uijlings, Koen~EA Van De~Sande, Theo Gevers, and Arnold~WM Smeulders.
\newblock Selective search for object recognition.
\newblock {\em International journal of computer vision}, 104(2):154--171,
  2013.

\bibitem{maxdeeplab}
Huiyu Wang, Yukun Zhu, Hartwig Adam, Alan Yuille, and Liang-Chieh Chen.
\newblock Max-deeplab: End-to-end panoptic segmentation with mask transformers.
\newblock In {\em Proceedings of the IEEE/CVF conference on computer vision and
  pattern recognition}, pages 5463--5474, 2021.

\bibitem{solov1}
Xinlong Wang, Tao Kong, Chunhua Shen, Yuning Jiang, and Lei Li.
\newblock Solo: Segmenting objects by locations.
\newblock In {\em European Conference on Computer Vision}, pages 649--665.
  Springer, 2020.

\bibitem{wang2022freesolo}
Xinlong Wang, Zhiding Yu, Shalini De~Mello, Jan Kautz, Anima Anandkumar,
  Chunhua Shen, and Jose~M Alvarez.
\newblock Freesolo: Learning to segment objects without annotations.
\newblock In {\em Proceedings of the IEEE/CVF Conference on Computer Vision and
  Pattern Recognition}, pages 14176--14186, 2022.

\bibitem{solov2}
Xinlong Wang, Rufeng Zhang, Tao Kong, Lei Li, and Chunhua Shen.
\newblock Solov2: Dynamic and fast instance segmentation.
\newblock {\em Advances in Neural information processing systems},
  33:17721--17732, 2020.

\bibitem{densecl}
Xinlong Wang, Rufeng Zhang, Chunhua Shen, Tao Kong, and Lei Li.
\newblock Dense contrastive learning for self-supervised visual pre-training.
\newblock In {\em Proceedings of the IEEE/CVF Conference on Computer Vision and
  Pattern Recognition}, pages 3024--3033, 2021.

\bibitem{zhang2021weakly}
Dingwen Zhang, Junwei Han, Gong Cheng, and Ming-Hsuan Yang.
\newblock Weakly supervised object localization and detection: A survey.
\newblock {\em IEEE transactions on pattern analysis and machine intelligence},
  44(9):5866--5885, 2021.

\bibitem{zhang2019synthesizing}
Dingwen Zhang, Junwei Han, Yu Zhang, and Dong Xu.
\newblock Synthesizing supervision for learning deep saliency network without
  human annotation.
\newblock {\em IEEE transactions on pattern analysis and machine intelligence},
  42(7):1755--1769, 2019.

\bibitem{zhang2020weakly}
Dingwen Zhang, Wenyuan Zeng, Jieru Yao, and Junwei Han.
\newblock Weakly supervised object detection using proposal-and semantic-level
  relationships.
\newblock {\em IEEE Transactions on Pattern Analysis and Machine Intelligence},
  2020.

\bibitem{knet}
Wenwei Zhang, Jiangmiao Pang, Kai Chen, and Chen~Change Loy.
\newblock K-net: Towards unified image segmentation.
\newblock {\em Advances in Neural Information Processing Systems},
  34:10326--10338, 2021.

\bibitem{zhao2021weakly}
Wangbo Zhao, Jing Zhang, Long Li, Nick Barnes, Nian Liu, and Junwei Han.
\newblock Weakly supervised video salient object detection.
\newblock In {\em Proceedings of the IEEE/CVF conference on computer vision and
  pattern recognition}, pages 16826--16835, 2021.

\bibitem{cooop}
Kaiyang Zhou, Jingkang Yang, Chen~Change Loy, and Ziwei Liu.
\newblock Conditional prompt learning for vision-language models.
\newblock In {\em Proceedings of the IEEE/CVF Conference on Computer Vision and
  Pattern Recognition}, pages 16816--16825, 2022.

\bibitem{coop}
Kaiyang Zhou, Jingkang Yang, Chen~Change Loy, and Ziwei Liu.
\newblock Learning to prompt for vision-language models.
\newblock {\em International Journal of Computer Vision}, 130(9):2337--2348,
  2022.

\bibitem{zhuge2022salient}
Mingchen Zhuge, Deng-Ping Fan, Nian Liu, Dingwen Zhang, Dong Xu, and Ling Shao.
\newblock Salient object detection via integrity learning.
\newblock {\em IEEE TPAMI}, 2022.

\end{thebibliography}
}

\end{document}